\documentclass{article}

\PassOptionsToPackage{numbers,compress,sort}{natbib}

\usepackage[final]{neurips_2024}

\usepackage[utf8]{inputenc} 
\usepackage[T1]{fontenc}    
\usepackage{hyperref}       
\usepackage{url}            
\usepackage{booktabs}       
\usepackage{amsfonts}       
\usepackage{nicefrac}       
\usepackage{microtype}      
\usepackage{xcolor}         
\usepackage{multirow}
\usepackage{xspace}
\usepackage{amsmath}
\usepackage[capitalise,noabbrev]{cleveref}
\usepackage{tabularx}
\usepackage{graphicx}
\usepackage{subcaption}
\usepackage{adjustbox}
\usepackage{listings}
\usepackage{enumitem}
\usepackage{float}
\usepackage{bbm}
\usepackage{wrapfig}
\usepackage{enumitem}
\usepackage{tikz}
\usetikzlibrary{arrows,calc,arrows.meta,positioning,shapes}
\usepackage{pdflscape}

\newcommand{\todo}[1]{\textcolor{red}{TODO: #1}}

\newcommand{\name}{{OLLM}\xspace}

\DeclareMathOperator{\nodesim}{NodeSim}

\hypersetup{
    colorlinks=true,
    linkcolor=blue,
    urlcolor=blue,
    citecolor=blue,
}

\title{End-to-End Ontology Learning with \\Large Language Models}

\author{%
    Andy Lo \\
    University of Cambridge\\
    \texttt{cyal4@cam.ac.uk} \\
    \And
    Albert Q. Jiang \\
    University of Cambridge\\
    \texttt{qj213@cam.ac.uk} \\
    \AND
    Wenda Li \\
    University of Edinburgh\\
    \texttt{wenda.li@ed.ac.uk} \\
    \And
    Mateja Jamnik \\
    University of Cambridge\\
    \texttt{mateja.jamnik@cl.cam.ac.uk} \\
}

\begin{document}

\maketitle

\begin{abstract}
Ontologies are useful for automatic machine processing of domain knowledge as they represent it in a structured format.
Yet, constructing ontologies requires substantial manual effort. To automate part of this process, large language models (LLMs) have been applied to solve various subtasks of ontology learning. However, this partial ontology learning does not capture the interactions between subtasks. 
We address this gap by introducing \name, a general and scalable method for building the taxonomic backbone of an ontology from scratch.
Rather than focusing on subtasks, like individual relations between entities, we model entire subcomponents of the target ontology by finetuning an LLM with a custom regulariser that reduces overfitting on high-frequency concepts. We introduce a novel suite of metrics for evaluating the quality of the generated ontology by measuring its semantic and structural similarity to the ground truth. In contrast to standard syntax-based metrics, our metrics use deep learning techniques to define more robust distance measures between graphs.
Both our quantitative and qualitative results on Wikipedia show that \name outperforms subtask composition methods, producing more semantically accurate ontologies while maintaining structural integrity. We further demonstrate that our model can be effectively adapted to new domains, like arXiv, needing only a small number of training examples. Our source code and datasets are available at \url{https://github.com/andylolu2/ollm}.

\end{abstract}

\section{Introduction}

\input{figures/overview_v6}

An ontology is a formal and structural way of representing domain-specific concepts and their relations~\cite{gruber1995toward}.
They can be simple (e.g., Wikipedia categories) consisting of \emph{concepts} and only a small number of types of \emph{taxonomic relations} (e.g., \emph{is-a} relationships), or they can be complex (e.g., Schema.org) consisting of axioms or many types of relations. For example, a simple ontology for programming languages might contain two concepts ``Dynamically-typed language'' and ``Python'', and one relation ``Dynamically-typed language $\to$ Python'', representing the knowledge that Python is a dynamically-typed language. A more complex ontology might contain axioms too, for example, ``all programming languages are either dynamically or statically typed''.
In this paper, we focus on ontologies with only concepts and taxonomic relations. Compared to typical deep learning models, which represent knowledge implicitly in its weights, ontologies capture knowledge in a structured and explicit manner, making them reliable, easy to edit and human-interpretable. Such benefits of ontologies have led to their wide adoption in practice. For example, Wikipedia categories have been used for entity ranking~\cite{vercoustre2008using} and information retrieval~\cite{sorg2012exploiting}, or Schema.org~\cite{Schema.org_2011} is a core component of the Semantic Web~\cite{antoniou2004semantic} initiative.

While ontologies are useful, building ontologies often requires substantial manual effort. Ontology learning (OL) is the study of automating the construction of high-quality ontologies at scale. For a simple ontology, this amounts to discovering the concepts and taxonomic relations, usually based on a source corpus. In this paper we aim to develop domain-independent methods for OL that are scalable and produce better ontologies.

Traditionally, OL is viewed as a composition of subtasks~\cite{asim2018survey}, such as concept discovery and relation extraction. In particular, prior works have demonstrated that state-of-the-art large language models (LLMs) can solve such subtasks effectively~\cite{babaei2023llms4ol}. While studying subtasks permits fine-grained analysis and evaluation, it does not directly indicate the subsequent impact on the quality of the final ontology. Moreover, there is potential room for improvement by combining several subtasks into one, such as by modelling concepts and relations in conjunction. In this paper, we instead develop and evaluate methods that construct ontologies in an end-to-end fashion to answer the following research questions:
\begin{enumerate}[itemsep=0pt,leftmargin=*]
    \item How can we leverage LLMs' knowledge base to build ontologies from scratch?
    \item Does our method scale efficiently to practical problem sizes?
    \item How well does our method generalise to new domains?
\end{enumerate}

We introduce \name, an end-to-end method for using LLMs to construct ontologies at scale. Rather than focusing on individual relations between concepts, we finetune an LLM to model entire sub-components of the target ontology. The output ontology is generated by taking the sum of generated sub-components and applying simple post-processing. An overview of the pipeline is shown in \cref{fig:overview}. To train \name, we collect the categorisation metadata for a subset of Wikipedia articles. We attempt to adapt an LLM to model the relevant categorisation subgraph for a particular Wikipedia article, but discover that direct finetuning leads to poor generalisation due to overfitting to high-level, frequently occurring concepts. Instead, we propose a custom regulariser that reweights each concept based on its frequency of occurrence, which substantially improves generalisation. 

We evaluate \name by measuring the similarity of the generated ontology with the ground truth. Current approaches for comparing ontologies rely on mapping components of the two ontologies onto each other, most commonly by literal text matching \cite{maedche2002measuring,Treeratpituk2013GraphbasedAT}. This is unreliable when the two ontologies are not already sufficiently similar. Instead, we propose a suite of evaluation metrics suitable for comparing arbitrary labelled graphs. These metrics compare edges and subgraphs of the two ontologies using pretrained text embedders to test for semantic and structural similarity. Both our quantitative and qualitative results reveal that an LLM can already outperform existing extraction-based methods out of the box, and the performance is further improved by finetuning with our custom regulariser. We additionally demonstrate that \name can be adapted to build the arXiv ontology using only a small number of training examples, suggesting that our model can be applied to new domains in a data-efficient way. In summary, our contributions are:
\begin{enumerate}[itemsep=0pt,leftmargin=*]
    \item We constructed two datasets based on Wikipedia and arXiv, which can serve as standard datasets for future work studying end-to-end OL.
    \item We created \name, a method that utilises LLMs to build ontologies from scratch. \name produces high-quality ontologies and serves as a strong baseline for end-to-end OL.
    \item We developed new evaluation metrics for assessing the quality of the generated ontologies. 
\end{enumerate}

\section{Background}

An ontology is a structured way of representing concepts and relations of a shared conceptualisation, that is, domain knowledge~\cite{gruber1995toward,gruber1993translation}. Ontologies can span a wide range of complexities. A fully-fledged ontology might contain concepts, relations, constraints, and axioms that enable complex automated reasoning. In this paper, we focus on the core building blocks of an ontology: concepts and taxonomic relations which represent \emph{is-a} or \emph{is-subclass-of} relationships between concepts. In some cases, the \emph{is-part-of} relation is also considered a taxonomic relation. We treat such an ontology as a rooted labelled directed graph where nodes represent concepts, edges represent taxonomic relations and the root node is the special concept of all concepts. A strict ontology asserts that the taxonomic relation is asymmetric and thus the graph must be acyclic, though in practice some ontologies, such as the Wikipedia ontology studied in this paper, may contain cycles. We therefore do not assume that an ontology graph is necessarily acyclic. Examples of ontologies include WordNet~\cite{miller1995wordnet} with 117,659 concepts and 89,089 taxonomic relations, and the Gene Ontology~\cite{ashburner2000gene} with 42,255 concepts and 66,810 taxonomic relations.

Ontology learning is the automatic extraction of ontological elements~\cite{hazman2011survey}. The most studied source of input is unstructured text, though there are also works on semi-structured data like HTML~\cite{karoui2004ontology}. In this paper, the input is a set of documents, each consisting of some unstructured text. We additionally assume each document is associated with one or more concepts in the ground truth ontology, which we utilise for training. The goal is to reconstruct the ground truth ontology given the set of documents.

Prior works view OL as a composition of subtasks, and study each subtask in isolation~\cite{buitelaar2005ontology,asim2018survey}. A typical pipeline for building a simple ontology is to first perform concept discovery (identify the nodes), and then relation extraction (identify the edges)~\cite{cimiano2005text2onto,kaushik2018automatic}. A notable approach for relation extraction is Hearst patterns~\cite{hearst1998automated}. Hearst patterns are hand-crafted lexico-syntactic patterns that exploit natural language structure to discover taxonomic relations. For example, the pattern ``[noun phrase] such as [noun phrase]'' matches phrases like ``dogs such as chihuahuas'', and thus can be processed by regular expressions to identify the relation ``dog $\to$ chihuahua''. Hearst patterns suffer from low recall, as the relations must occur in exact configurations to be identified by the rules. \citet{roller2018hearst} suggest smoothing techniques to alleviate this issue though at the cost of lower precision.

Recently, language models have been used for OL. REBEL~\cite{cabot2021rebel} treats relation discovery as a translation task, and finetunes encoder-decoder LLMs to extract both taxonomic and non-taxonomic relations. \citet{babaei2023llms4ol} benchmarked a wide family of LLMs for concept and relation discovery, and showed promising results. However, the quadratic complexity of link prediction makes this approach unscalable to large ontologies. We provide more discussion in \cref{appendix:llms4ol}. There are also proof-of-concept works for building ontologies end-to-end with LLMs. \citet{funk2023towards} proposes to build an ontology by recursively prompting  LLMs, while \citet{trajanoska2023enhancing} generate the entire ontology in one completion. However, both studies are limited in the scale of the task and evaluation: they only considered ontologies of up to 1000 concepts and relied on manual qualitative evaluation. We bridge this gap by proposing a method that can scale to practical problem sizes and new metrics for systematic qualitative evaluation.

The evaluation of ontologies is an open research area. The main approaches are gold standard evaluation~\cite{Zavitsanos2011GoldSE}, which matches elements of the generated ontology with a predefined target ontology; task-based evaluation~\cite{porzel2004task}, which measures the usefulness of the ontology on a specific application; and human evaluation \cite{raad2015survey,brank2005survey}. In this paper, we evaluate by the gold standard metric as it is the most straightforward approach when ground-truth ontology exists. Prior works have considered matching concepts~\cite{maedche2002measuring} and direct or indirect relations~\cite{Kashyap2005TaxaMinerAE, Treeratpituk2013GraphbasedAT} by literal text comparison. Others have also considered edit-distance~\cite{Ehrig2005SimilarityFO} or bag-of-words distributional similarity for text comparison~\cite{Zavitsanos2011GoldSE}.  These techniques for measuring semantic similarity may be considered unreliable and have been superseded by current methods~\cite{conneau2017supervised}. We instead rely on more modern techniques like pretrained text embedders \cite{devlin2018bert} and graph convolutions \cite{kipf2016semi} to match substructures between the two ontologies. 

\section{\name}

We now introduce \name, our novel, simple and scalable method for end-to-end OL with LLMs. On a high level, \name uses an LLM to model concept subgraphs of the target ontology by utilising a linearisation scheme to transform subgraphs into string sequences. In contrast to learning individual edges, modelling subgraphs allows the model to learn higher-order structures, such as the interactions between three or more nodes. To create the training dataset, \name relies on the annotations of documents to concepts to generate document-subgraph pairings. Such subgraphs are much smaller than the complete graph, so they can be learned by the model more easily. The generated subgraphs for each document are summed into a weighted graph, and simple post-processing is applied to obtain the final predicted ontology.

\subsection{Subgraph modelling}  \label{sec:method:subgraph}

\begin{figure}[t]
    \centering
    \begin{subfigure}[c]{0.335\textwidth}
        \centering
        \fbox{
        \includegraphics[width=\linewidth,trim={1.5cm 1.5cm 1.5cm 1.5cm},clip]{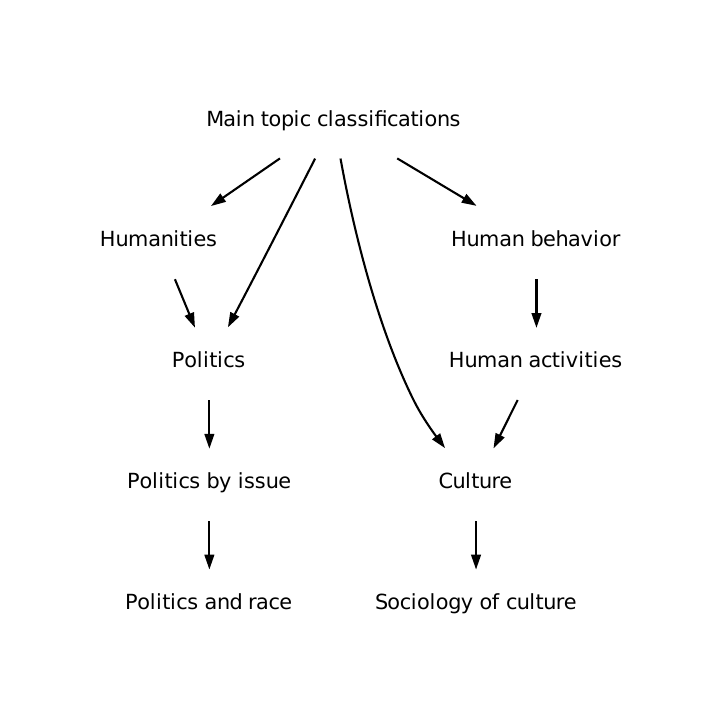}
        }
    \end{subfigure}%
    \hfill
    \begin{adjustbox}{varwidth=\linewidth,fbox}
    \begin{subfigure}[c]{0.55\textwidth}
        \centering
        \lstdefinestyle{prompt}{
          basicstyle=\small\ttfamily\color{black},
          moredelim=**[is][\color{gray}]{@}{@},
        }
        \begin{lstlisting}[gobble=8,style=prompt]
        @<s>[INST] Title: Hybridity
        Hybridity, in its most basic sense ... [/INST]@
        Main topic classifications -> @Human behavior@ -> @Human activities@ -> Culture -> Sociology of culture
        Main topic classifications -> @Humanities@ -> Politics -> @Politics by issue@ -> @Politics and race@
        Main topic classifications -> @Politics@ -> @Politics by issue@ -> Politics and race
        Main topic classifications -> @Culture@ -> Sociology of culture</s>
        \end{lstlisting}
    \end{subfigure}
    \end{adjustbox}
    \caption{Example subgraph induced for the Wikipedia page ``Hybridity'' (left), where $N = 4$ and $C = \{\text{Politics and race}, \text{Sociology of culture}\}$.
    The corresponding training text sequence (right), where text coloured in grey is ignored as training targets, but is still present as context for later tokens.}
    \label{fig:prompt-example}
\end{figure}

Here, we describe the method for creating document-subgraph pairings. Given a document and its associated set of concepts $C$, we define the \emph{relevant paths} as the set of paths of at most length $N$ from the root to any of the concepts in $C$. The \emph{relevant subgraph} is the set of nodes (concepts) and edges (taxonomic relations) that occur at least once in the relevant paths. An example is shown in \cref{fig:prompt-example} (left). The choice of $N$ is task-specific and we describe our method for choosing $N$ in \cref{sec:implementation}. 

To employ LLMs to model the subgraphs, we must linearise the graph into a string sequence. Existing methods for autoregressive graph generation employ BFS \cite{you2018graphrnn} or DFS \cite{goyal2020graphgen} ordering starting at an arbitrary node. We instead choose to linearise the subgraph as a list of relevant paths that produced the subgraph in the first place. We do so over BFS/DFS ordering for three reasons: 1)~the subgraph is defined from the relevant paths, which makes them the most natural representation; 2)~we hypothesise that the hierarchy of concepts in each path is a desirable inductive bias for the hierarchical nature of an ontology; and 3)~the path-based representation is much easier to describe in natural language instructions so that our LLM prompting-based baselines may produce reasonable results without finetuning. The linearisation template can be found in \cref{fig:linearisation-template} in Appendix~\ref{appendix:training-details}.

\subsection{Post-processing}  \label{sec:method:post-processing}

The final output graph is obtained by summing all generated subgraphs for each document and pruning low-weighted components. Given the generated subgraphs $G_1 = (V_1, E_1), \dots, G_n = (V_n, E_n)$, the raw output graph is defined as $G_\text{raw} = (V_\text{raw}, E_\text{raw})$, where $V_\text{raw} = \cup_{i=1}^n V_n$ and $E_\text{raw} = \cup_{i=1}^n E_n$. Each edge $(u, v) \in E_\text{raw}$ is additionally weighted by the number of times it occurs in the collection of subgraphs: $w(u, v) = \sum_{i=1}^n \mathbbm{1}[(u,v) \in E_n]$. A few simple post-processing steps are then applied to $G_\text{raw}$ in order to prune it:
\begin{enumerate}[itemsep=0pt,leftmargin=*]
    \item Self-loop pruning: All edges $(u, u) \in E_\text{raw}$ are removed.
    \item Inverse-edge pruning: For $(u, v) \in E_\text{raw}$, if $(v, u) \in E_\text{raw}$ and $w(v, u) > w(u, v)$, remove $(u, v)$. That is, bidirectional edges are turned into unidirectional ones.
    \item Absolute thresholding: Edges in $E_\text{raw}$ with weight below the $\alpha$-th quantile are removed, where $0 \leq \alpha \leq 1$ is a hyperparameter. This removes edges that are globally less important.
    \item Relative thresholding: For each vertex $u \in V_\text{raw}$, let $e_1, \dots, e_k$ be the outgoing edges from $u$, sorted by weight in ascending order. Let the cumulative weight be $C(e_i) = \sum_{j=1}^i w(e_j) / \sum_{j=1}^k w(e_j)$. The edges $\{e_i\ |\ C(e_i) \leq \beta\}$ are pruned, where $0 \leq \beta \leq 1$ is a hyperparameter. This is similar to top-$p$ sampling \cite{holtzman2019curious}, which we use to remove edges that are less important than their neighbours.
    \item Clean up: After pruning all edges, nodes with no incoming or outgoing edges are removed.
\end{enumerate}
We choose the hyperparameters $\alpha$ and $\beta$ by tuning on the validation set (\cref{sec:implementation}).

\section{Evaluating end-to-end OL}

Ontology evaluation is a hard problem as there are no quantitative definitions of what constitutes a ``good ontology'', and metrics generally only capture one aspect (e.g., structure but not semantics) of an ontology. We approach evaluation by treating the ground truth as a proxy for a good ontology, and comparing the generated ontologies against the ground truth. Here, we describe how the ground truth is obtained, and introduce new evaluation metrics that are used for measuring ontology similarity.

\subsection{Dataset}  \label{sec:dataset}

We collect the datasets for the two ontologies considered in this paper: Wikipedia categories and the arXiv taxonomy. We use Wikipedia for learning and in-domain evaluation, and arXiv for out-of-domain evaluation. To build the Wikipedia dataset, we perform a BFS traversal from its root category ``Main topic classifications'' up to depth 3. For every category encountered, we retrieve the titles and summaries (the text before the first section) of up to 5000 pages that belong in that category. The source data is obtained from the Wikipedia API.\footnote{\url{https://en.wikipedia.org/w/api.php}} The arXiv taxonomy is available from its home page, and the source corpus is constructed from the title and abstract of all the papers uploaded to arXiv in the years 2020--2022 with more than or equal to 10 citations.\footnote{Citation counts obtained from \url{https://api.semanticscholar.org/}.} In total, the Wikipedia dataset has 13886 concepts, 28375 taxonomic relations and 362067 documents, while the arXiv dataset has 161 concepts, 166 taxonomic relations and 126001 documents.

\begin{wrapfigure}{r}{0.5\textwidth}
    \vspace{-3mm}
    \centering
    \begin{subfigure}[t]{0.5\linewidth}
        \centering
        \includegraphics[width=\linewidth,trim={0.4cm 0.65cm 0.5cm 0.9cm},clip]{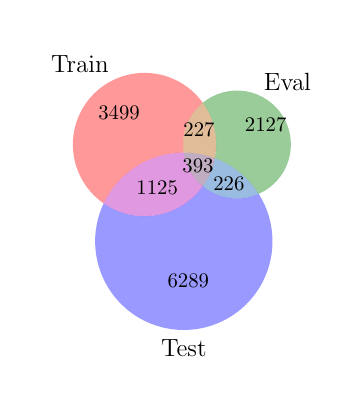}
        \caption{Wikipedia}
    \end{subfigure}%
    \begin{subfigure}[t]{0.4\linewidth}
        \centering
        \includegraphics[width=\linewidth,trim={1cm 1cm 1cm 1.2cm},clip]{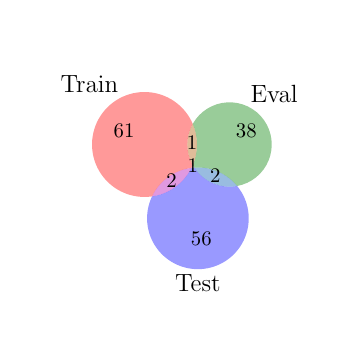}
        \caption{arXiv}
    \end{subfigure}
    \vspace*{-2mm}
    \caption{Intersection of concepts among the train, validation and test splits of the datasets.}
    \vspace*{-3mm}
    \label{fig:dataset-overlap}
\end{wrapfigure}

Generating the train and test splits from the datasets is a non-trivial problem. Each training example consists of a document and its relevant subgraph (\cref{sec:method:subgraph}). The naive approach of randomly selecting a subset of document-subgraph pairs for the training likely leads to data leakage as there might be a significant overlap between subgraphs in the training set and the test set. Instead, we first split the full ontology into train and test graphs, and then generate the training document-subgraph pairs. This ensures that there are sufficiently many unseen concepts (and thus relations) in the test split, as shown in \cref{fig:dataset-overlap}. Our method is as follows:
\begin{enumerate}[leftmargin=*]
    \item Let $V^\text{top}$ be the set of top-level nodes, that is, children of the root node. Randomly partition $V^\text{top}$ into train $V^\text{top}_{\text{train}}$, validation $V^\text{top}_{\text{val}}$, and test $V^\text{top}_{\text{test}}$ splits in 7:3:10 ratio.
    \item Let $d$ be the depth on the full graph, that is, the distance of the furthest node from the root. The nodes of the train graph are taken as the union of all the nodes that are within distance $d - 1$ from any node in $V^\text{top}_\text{train}$, plus $V_\text{train}^\text{top}$ and the root. The edges are all the edges in the full graph that have both endpoints in the train graph. Similar applies for $V^\text{top}_\text{val}$ and $V^\text{top}_\text{test}$.
\end{enumerate}

\subsection{Metrics}

Many existing methods for comparing ontologies rely on syntactic measures like string edit distance~\cite{Ehrig2005SimilarityFO} as a proxy for semantic similarity, or require every concept to be tagged with descriptions or documents for distributional semantics comparison~\cite{Zavitsanos2011GoldSE}. To obtain more robust and general evaluation results, we introduce a suite of similarity metrics that use modern methods like text embeddings~\cite{reimers-2019-sentence-bert}. Multiple metrics are used as they trade off between interpretability and comprehensiveness and we aim to make them complementary by capturing different aspects of an ontology. For example, comparing ontologies by literal text equality is easy to understand but may be unreliable. In \cref{sec:meta-eval}, we provide further discussion on evaluation metrics in the context of our experiment results. We denote the ground truth ontology graph as $G = (V, E)$ and the generated graph as $G' = (V', E')$. 

\textbf{Literal~F1 }
While literal text matching is unreliable, it is also the simplest and the most interpretable. We treat this metric as a reference metric for sanity check. The Literal~F1 metric~\cite{Kashyap2005TaxaMinerAE} is given by the harmonic mean of the precision and recall of the edges:
\[
\text{Literal precision} = \frac{|E \cap E'|}{|E'|} \qquad
\text{Literal recall} = \frac{|E \cap E'|}{|E|}
\]

\textbf{Fuzzy~F1 }
The Literal~F1 metric puts a strong emphasis on using the correct wording, while in practice, we are interested in evaluating the semantics of an ontology. For example, using a synonymous phrase for a concept should not be penalised. We utilise embeddings from a pretrained sentence transformer~\cite{reimers-2019-sentence-bert} and use the cosine similarity of the embeddings to measure semantic similarity. Specifically, let $\nodesim(u, u') \in V \times V' \to [-1, 1]$ be the cosine similarity between the sentence embeddings for $u$ and $u'$. The Fuzzy~F1 score is obtained from the fuzzy precision and recall, defined as:\looseness-1
\begin{equation*}
\begin{aligned}
\text{Fuzzy precision} &= \frac{|
\{(u', v') \in E' \mid \exists (u, v) \in E. 
\nodesim(u, u') > t \land \nodesim(v, v') > t
\}
|}{|E'|} \\
\text{Fuzzy recall} &= \frac{|
\{(u, v) \in E \mid \exists (u', v') \in E'. 
\nodesim(u, u') > t \land \nodesim(v, v') > t
\}
|}{|E|} \\
\end{aligned}
\end{equation*}
where $t$ is the matching threshold. We use all-MiniLM-L6-v2~\cite{wang2020minilm,reimers-2019-sentence-bert} as the embedding model, and choose $t$ as the median cosine similarity between the synonyms in WordNet~\cite{miller1995wordnet}, computed as~0.436.

\textbf{Continuous~F1 }
With fuzzy comparisons, the matches between the edges of the generated and the ground truth graph are no longer one-to-one. This is problematic: consider two graphs $A\!\rightarrow\!B$ and $B\!\leftarrow\!A\!\rightarrow\!B'$, where $B$ and $B'$ match fuzzily. Such graphs will achieve a perfect Fuzzy~F1 score yet they significantly differ. Additionally, we found that the previous metrics fail to provide a useful signal for hyperparameter tuning, particularly for our baselines where the generated graphs are poor. The Continuous~F1 metric solves these issues by computing the highest-scoring edge matching between the two graphs, where the similarity score between $(u, v)$ and $(u', v')$ is given by $\min(\nodesim(u, u'), \nodesim(v, v'))$. Obtaining such matching is equivalent to solving the linear assignment problem~\cite{martello1987linear}, which can be computed by the Hungarian algorithm~\cite{kuhn1955hungarian}. The Continuous~F1 score is obtained from the continuous precision and recall, given by:
\[
\text{Continuous precision} = \frac{s_\text{cont}}{|E'|} \qquad
\text{Continuous recall} = \frac{s_\text{cont}}{|E|}
\]
where $s_\text{cont}$ is the score achieved by the best edge matching.

\textbf{Graph~F1 }
Instead of individual edges, this metric aims to capture the wider structure of the two graphs. Intuitively, we want to know how concepts are related to their local neighbourhood. We do so by using simple graph convolutions~\cite{wu2019simplifying} with $K=2$ to compute graph-aware node embeddings after embedding each node with the pretrained embedder. Such embeddings in $G$ are compared against those in $G'$ by cosine similarity, and the highest-scoring node matching, similar to the Continuous~F1 metric, gives the graph similarity score. The Graph~F1 score is computed from the graph precision and recall, defined as:
\begin{samepage}
\[
\text{Graph precision} = \frac{s_\text{graph}}{|V'|} \qquad
\text{Graph recall} = \frac{s_\text{graph}}{|V|}
\]
where $s_\text{graph}$ is the score achieved by the best node matching.
\end{samepage}

\textbf{Motif distance }
Taking inspiration from classical network analysis, we use \emph{network motifs}~\cite{milo2002network,shen2002network} to evaluate the structural integrity of the generated graphs. Network motifs are reoccurring subgraphs in a larger graph, most commonly 3-vertex subgraphs. They are typically indicative of the structural characteristics of the full graph. We define the motif distance as the total variation distance between the distribution of all 3-vertex subgraphs in $G$ and $G'$.

\section{Experiments}

We design our experiments to answer the following research questions:
\begin{enumerate}[leftmargin=*]
    \item Does \name produce better ontologies than traditional methods by subtask composition?
    \item Can \name be easily adapted to a new domain?
\end{enumerate}
 We approach the questions by training \name on the Wikipedia dataset, and further transfer the model to arXiv with a small number of arXiv samples. As baselines, we use two relation extraction methods, Hearst patterns~\cite{hearst1998automated,roller2018hearst} and REBEL~\cite{cabot2021rebel}. Relation extraction depends on successful concept discovery to produce high-quality ontologies. To estimate a ceiling to such baselines, \emph{we give the baselines a substantial advantage} by providing them with the ground truth concepts in the test graph. The results show that even with such an advantage, \name outperforms the baselines on many metrics, demonstrating the potential of \name for end-to-end OL (\cref{sec:results}).

\subsection{Implementation details}  \label{sec:implementation}

\begin{figure}[t]
\vspace*{-2mm}  
    \centering
    \begin{subfigure}[c]{\textwidth}
        \centering
        \includegraphics[width=\linewidth,trim={0 1cm 0 0},clip]{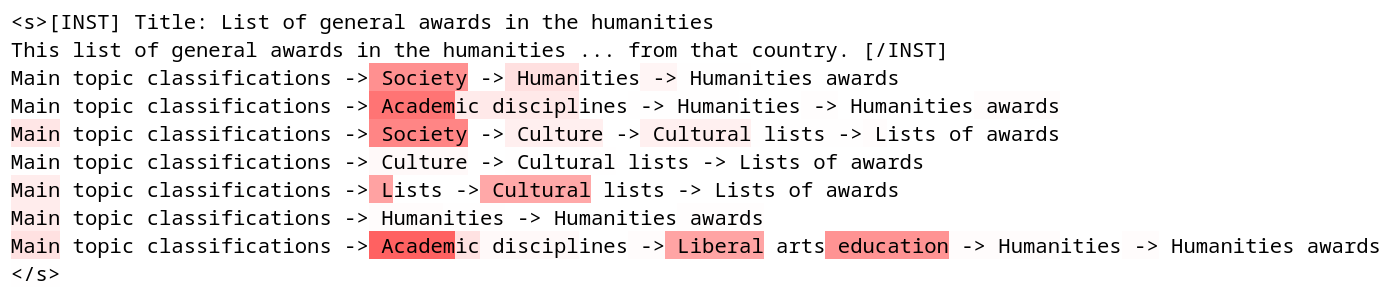}
        
        \vspace*{-1mm}
        \caption{Direct finetuning}
    \end{subfigure}%
    \hfill
    \begin{subfigure}[c]{\textwidth}
        \centering
        \includegraphics[width=\linewidth,trim={0 1cm 0 0},clip]{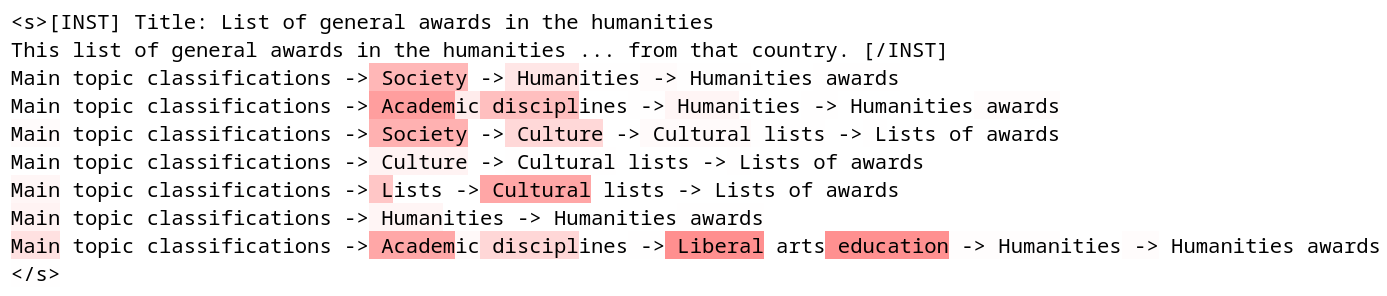}
        
        \vspace*{-1mm}  
        \caption{Finetuning with masked loss}
    \end{subfigure}%
    \caption{Per token loss on a test set example of the final model trained with and without the custom masked loss objective. A stronger red colour represents a higher cross-entropy loss. Within the top-level concepts (children of the root) shown here, ``Culture'' and ``Humanities'' are in the training set while others are not. Using the masked loss objective improves generalisation on the high-level relations (e.g., ``Main topic classifications'' $\rightarrow$ ``Academic disciplines'') while maintaining performance on lower-level relations.}
    \label{fig:vanilla-vs-mask}
    \vspace*{-2mm}  
\end{figure}

Analysing the per-token loss on the test split sequences of a directly finetuned model (\cref{sec:method:subgraph}) shows that the model tends to memorise high-level relations from the training set, leading to poor generalisation, as shown in \cref{fig:vanilla-vs-mask} (top).

The crux of the problem is that low-level relations are substantially more diverse than high-level ones: since we present both types of relations at the same rate to the model, it tends to overfit on high-level relations while underfitting on low-level ones. 
To alleviate this issue, we introduce a new training objective that randomly masks the loss contribution of frequently occurring relations. Suppose a relation $u \to v$ is present $n$ times in the training set. During training, when $u \to v$ appears in one of the relevant paths, we mask the loss contribution of the tokens for $v$ with probability $\max(1 - \nicefrac{M}{n}, 0)$, where $M$ is a constant for the average number of times a relation is present in the training set. Intuitively, this regulariser ensures that frequent relations are only seen $\approx\!M$ times as targets throughout training, hence reducing overfitting as shown in \cref{fig:vanilla-vs-mask} (bottom). Note that while $v$ is masked from the target, its tokens are still present in the input sequence as context for later tokens. A concrete training example can be found in \cref{fig:prompt-example} (right).\looseness-1 

We finetune Mistral 7B v0.2~\cite{jiang2023mistral} with Low-Rank Adaptation~\cite{hu2021lora} on the masked loss objective. The model is trained on the Wikipedia dataset for two epochs with Adam~\cite{kingma2014adam}. During inference, the outputs are generated with temperature 0.1 and nucleus sampling~\cite{holtzman2019curious} top-$p$ of 0.9. We include a finetuning baseline without the masked loss objective, denoted as Finetune. To adapt \name for arXiv, we further finetune the model on 2048 document-subgraph pairs from arXiv. We initialise new low-rank adaptors and train until the loss stops improving on the validation set. We name these models \name (transfer) and Finetune (transfer) for training with and without the masked loss objective, respectively. Full details for the Wikipedia and arXiv experiments can be found in~\cref{appendix:training-details}.

The hyperparameters for the post-processing steps are tuned by grid search on the validation set. We sweep over $\alpha \in 1 - \text{geomspace}(1 / |E_\text{raw}|, 1, 21)$ and $\beta \in \text{geomspace}(0.1, 1, 21) - 0.1$, and use the values that maximise Continuous~F1. For Wikipedia, we choose the subgraph modelling path length $N=4$ as it is the smallest $N$ such that almost all edges ($>99\%$) occur in at least one relevant subgraph. Such criterion is used since smaller $N$ results in smaller subgraphs, which we expect to be easier to model accurately. We choose $N=3$ for arXiv for the same reason. 

\subsection{Baselines}

We give a brief overview of the baseline methods here (in addition to Finetune and Finetune (transfer)). The full implementation details can be found in \cref{appendix:exp-details}. All baselines produce weighted directed graphs which we apply the same post-processing steps as in \name (\cref{sec:method:post-processing}) to obtain the final predicted graph.

\textbf{Memorisation }
Simply memorising the train graph is a surprisingly strong baseline due to the overlap between train and test graphs, especially for Wikipedia. The weight of each edge is given by the number of relevant subgraphs in which it appears.

\textbf{Hearst }
We follow the improved implementation of Hearst patterns by \citet{roller2018hearst}. The authors propose spmi, a method which uses low-rank approximations to smooth the relation matrix so that two concepts can be compared even if there are no direct matches between them. We use the smoothed relation matrix to weigh the relations between the ground truth concepts. The additional hyperparameter for the rank of the smoothed matrix is tuned by grid search over the validation set.

\textbf{REBEL }
The REBEL-large model~\cite{cabot2021rebel} is an LLM trained to extract many types of relations from Wikipedia articles. We only take the ``subclass of'', ``instance of'', ``member of'' and ``part of'' relations that were extracted. Similar to Hearst, we find that it fails to find many direct relations between ground truth concepts. The same low-rank smoothing technique is applied to improve recall. 

\textbf{Prompting }
We test the Zero/One/Three-shot performance of instruction-tuned LLMs on the subgraph modelling task described in \cref{sec:method:subgraph}. To obtain more comparable results, we use Mistral 7B Instruct v0.2, the instruction-tuned version of the base model of \name, as the LLM for our prompting baseline. The prompt template used is shown in \cref{fig:prompt-template} in Appendix~\ref{appendix:exp-details}.

\textbf{Finetune }
To test the effectiveness of our masked-loss objective, we introduce a direct finetuning baseline using the same configuration as \name except it is trained without loss masking.

\subsection{Results}  \label{sec:results}

We first evaluate whether \name can accurately create ontologies with many concepts and relations, such as the Wikipedia categories. Computationally, \name required 12 A100-hours for training and 7 A100-hours for inference to generate an ontology for Wikipedia. This is a modest cost in current standards, which demonstrates the scalability of \name for real-world problems. 
In terms of performance, \name produces the most semantically accurate ontology in comparison to our baselines as presented in \cref{table:metrics}. Across all of Fuzzy~F1, Continuous~F1 and Graph~F1, we observe the trend that \name scores the best, followed by Finetune and Prompting, and lastly Hearst and REBEL. This is surprising, as it suggests that the combination of LLMs with our subgraph modelling framework is a sufficiently strong inductive bias for LLMs to outperform traditional methods even without finetuning. However, prompting alone is not sufficient to build high-quality ontologies. 
On the Motif Distance metric, prompting methods score poorly at 0.314--0.354 in comparison to 0.050 and 0.080 for Finetune and \name respectively. This shows that using LLMs out-of-the-box for subgraph modelling results in poor structural integrity, though this issue is solved by finetuning. 
Qualitatively, we observe that \name can adhere to the clear, explicit naming style of Wikipedia, even on unseen topics in the test set. For example, it generates ``Mathematical categories'' and ``Groups (mathematics)'' under the parent concept ``Mathematical structures'' to distinguish from the natural language sense of categories and groups (\cref{fig:ollm-wiki-samples-math}). Such style is not learned by the prompting baselines: Three-shot generated ``Elections $\to$ France'', while it most likely meant ``Elections $\to$ Elections in France'' (\cref{fig:3shot-wiki-samples-election}). More sample outputs are shown in \cref{appendix:viz-wiki}.

{
\addtolength{\tabcolsep}{-0.2em}
\begin{table}[t!]
\caption{Evaluation metrics of \name and baselines on Wikipedia and arXiv. \name performs particularly well in modelling semantics, and remains competitive syntactically and structurally.}
\label{table:metrics}
\centering
\begin{small}
\begin{tabularx}{\linewidth}{l l X X X X l}
\toprule
Dataset & Method & Literal F1 $\uparrow$ & Fuzzy F1 $\uparrow$ & Cont. F1 $\uparrow$ & Graph F1 $\uparrow$ & Motif Dist. $\downarrow$ \\
\midrule
\multirow[t]{8}{*}{Wikipedia} & Memorisation & \textbf{0.134} & 0.837 & 0.314 & 0.419 & 0.063 \\
 & Hearst & 0.003 & 0.538 & 0.350 & 0.544 & 0.163 \\
 & Rebel & 0.004 & 0.624 & 0.356 & 0.072 & 0.132 \\
 & Zero-shot & 0.007 & 0.871 & 0.455 & 0.639 & 0.341 \\
 & One-shot & 0.031 & 0.888 & 0.477 & 0.610 & 0.314 \\
 & Three-shot & 0.031 & 0.880 & 0.475 & 0.622 & 0.354 \\
 & Finetune & 0.124 & 0.884 & 0.470 & 0.588 & \textbf{0.050} \\
 & \textbf{\name} & 0.093 & \textbf{0.915} & \textbf{0.500} & \textbf{0.644} & 0.080 \\
\midrule
\multirow[t]{8}{*}{arXiv} & Memorisation & 0.000 & 0.207 & 0.257 & 0.525 & \textbf{0.037} \\
 & Hearst & 0.000 & 0.000 & 0.151 & 0.553 & 0.098 \\
 & Rebel & 0.000 & 0.060 & 0.281 & 0.546 & 0.088 \\
 & Zero-shot & 0.025 & 0.450 & 0.237 & 0.414 & 0.145 \\
 & One-shot & \textbf{0.072} & 0.460 & 0.290 & 0.433 & 0.293 \\
 & Three-shot & 0.051 & 0.405 & 0.212 & 0.385 & 0.124 \\
 & Finetune (transfer) & 0.000 & 0.440 & 0.225 & 0.441 & 0.148 \\
& \textbf{\name} (transfer) & 0.040 & \textbf{0.570} & \textbf{0.357} & \textbf{0.633} & 0.097 \\
\bottomrule
\end{tabularx}
\vspace*{-4mm}
\end{small}
\end{table}
}

The arXiv task differs from the Wikipedia task as it has much fewer relations, and there is even less overlap between the train and test split. This imposes a great challenge on Finetune and \name as they need to generalise with a limited diversity of training samples. Despite such constraints, \name is substantially better than other methods in modelling the semantics of the test graph. 
On the Fuzzy~F1, Continuous~F1, and Graph~F1 metrics, \name performs the best among all methods with 0.570, 0.357, and 0.633, significantly higher than the next-best of 0.460, 0.290 and 0.546 respectively. 
Inspecting the generated ontologies (\cref{appendix:viz-arxiv}), we observe that prompting baselines tend to produce repetitive concepts such as ``Machine Learning and Artificial Intelligence'' and ``Artificial Intelligence and Machine Learning'' (\cref{fig:3shot-arxiv}), while Hearst and REBEL put almost all concepts under the same parent concept(s) (\cref{fig:hearst-arxiv,fig:rebel-arxiv}). 
We also found that \name's output for arXiv contains concepts from Wikipedia, but restructured in a way that fits the arXiv ontology. For example, ``Life sciences'' and ``Biological evolution'' appear in the Wikipedia training set under the same parent category ``Life'' with no direct links between them. On the generated graph for arXiv, ``Life sciences'' is instead promoted to one of the top-level concepts with ``Biological Evolution'' as one of its children, which better fits the ``fields of science'' style of the arXiv ontology (\cref{fig:ollm-arxiv}). This demonstrates that \name can adapt to produce a new type of ontology by restructuring its learned concepts, all using just a small number of training samples. 

In summary, \name scores the best or is competitive across all metrics in both tasks, with the notable exception of the Literal~F1 metric. We attribute this to the fact that Literal~F1 is sensitive to factors like casing and choice of words, and generally only measures syntactic similarity. For example, we see that a suboptimal baseline like Memorisation scores the best on this metric with 0.134 on the Wikipedia task. This reflects that syntactic similarity generally does not entail semantic similarity, so syntax-based metrics should not be used as stand-alone measures for ontology quality. 

\subsection{Meta-evaluation}
\label{sec:meta-eval}

In this section, we analyse the usefulness of our new metrics for measuring graph similarity and discuss the limitations of existing metrics. On the Wikipedia task, Memorisation, despite being clearly the worst in Continuous F1 and Graph F1, performs the best on Literal F1 and the second-best on Motif Distance. This can be attributed to the fact that Literal F1 is sensitive to semantically insignificant syntactic differences such as casing and word form, and thus when the training and test set has non-trivial overlap (\cref{fig:dataset-overlap}), it is biased towards methods that overfit. Similarly, as per the method described in \cref{sec:dataset}, the data splits are constructed with structural symmetry, hence we expect the train and test splits to have a similar graph structure even though the represented concepts are different. As a result, methods that tend to overfit, for example, Memorisation and Finetune, achieve the best scores on Motif Distance. This demonstrates that Literal F1 and Motif Distance only capture syntactic and structural similarity respectively, and thus should not be used as stand-alone metrics for evaluation.

Analysing the edge and node matchings found by our Continuous F1 and Graph F1 metrics on arXiv reveals that they successfully capture some human intuition on semantic similarity between the two ontologies. In \cref{fig:edge-matching,fig:graph-matching}, we visualise the ontology generated by \name and the ground truth and observe that semantically similar components in the two graphs indeed get matched. For example, the ``Physics'' and ``Mathematics'' clusters in the generated graph get matched with the ``Mathematics'' cluster in the ground truth, ``Data Analysis'' and ``Information'' get matched with ``Statistics'', ``Economics'' with ``Quantitative Finance'', and ``Life Sciences'' with ``Quantitative Biology''. This suggests that our edge/node matching procedure is capturing a ``semantic graph isomorphism'' that allows one to compare similar components in the two graphs, even if they do not exactly share the same concepts. We believe this example of a semantic mapping from one ontology to another is strong evidence that our metrics are capturing meaningful qualities of the ontologies.\looseness-1

\section{Discussion}  \label{sec:disccusion}

\textbf{Limitations } We only study and evaluate the construction of simple ontologies with only concepts and taxonomic relations. A potential approach to extend \name to produce non-taxonomic relations is to add tags indicating the relation type to each edge when linearising the subgraphs for sequence modelling. New evaluation metrics might also be required to handle multiple types of relations. Another limitation is that the taxonomic relations in the generated ontologies are not necessarily transitive due to the existence of cycles. This is a general problem for many OL methods and there are existing works on cycle removal algorithms for cleaning hierarchies~\cite{sun2017breaking,zesch2007analysis}. We ablate this in \cref{appendix:consistency} and found that the generated ontology can be made consistent by removing a small number of edges.
Furthermore, we were unable to fully control for data contamination as the pretraining dataset of Mistral 7B is not publically known. We do, however, observe that the generated ontologies are sufficiently different from the ground truth, indicating that \name is not directly remembering samples from its pretraining stage.

\textbf{Conclusion } In this paper, we introduce a general method for building ontologies in an end-to-end fashion. We propose a set of metrics for end-to-end OL that measures the semantic and structural similarity between arbitrary labelled graphs. Our model, \name, outperforms traditional subtask composition methods in reconstructing the Wikipedia categories, and can be transferred to build ontologies for arXiv after finetuning on a small number of examples. Using LLMs as the backbone for subgraph modelling opens up exciting avenues for future research. For example, one may generate ontologies from corpora with images using vision language models~\cite{donahue2015long}.

\section{Acknowledgements}

We thank Dr Thomas Sauerwald for suggesting network motifs as a basis for evaluation. AQJ acknowledges the support of a Peterhouse Graduate Studentship.
 
\bibliographystyle{plainnat}
\bibliography{references}

\begin{thebibliography}{52}
\providecommand{\natexlab}[1]{#1}
\providecommand{\url}[1]{\texttt{#1}}
\expandafter\ifx\csname urlstyle\endcsname\relax
  \providecommand{\doi}[1]{doi: #1}\else
  \providecommand{\doi}{doi: \begingroup \urlstyle{rm}\Url}\fi

\bibitem[Antoniou and Van~Harmelen(2004)]{antoniou2004semantic}
Grigoris Antoniou and Frank Van~Harmelen.
\newblock \emph{A semantic web primer}.
\newblock MIT press, 2004.

\bibitem[Ashburner et~al.(2000)Ashburner, Ball, Blake, Botstein, Butler, Cherry, Davis, Dolinski, Dwight, Eppig, et~al.]{ashburner2000gene}
Michael Ashburner, Catherine~A Ball, Judith~A Blake, David Botstein, Heather Butler, J~Michael Cherry, Allan~P Davis, Kara Dolinski, Selina~S Dwight, Janan~T Eppig, et~al.
\newblock Gene ontology: tool for the unification of biology.
\newblock \emph{Nature genetics}, 25\penalty0 (1):\penalty0 25--29, 2000.

\bibitem[Asim et~al.(2018)Asim, Wasim, Khan, Mahmood, and Abbasi]{asim2018survey}
Muhammad~Nabeel Asim, Muhammad Wasim, Muhammad Usman~Ghani Khan, Waqar Mahmood, and Hafiza~Mahnoor Abbasi.
\newblock A survey of ontology learning techniques and applications.
\newblock \emph{Database}, 2018:\penalty0 bay101, 2018.

\bibitem[Babaei~Giglou et~al.(2023)Babaei~Giglou, D’Souza, and Auer]{babaei2023llms4ol}
Hamed Babaei~Giglou, Jennifer D’Souza, and S{\"o}ren Auer.
\newblock Llms4ol: Large language models for ontology learning.
\newblock In \emph{International Semantic Web Conference}, pages 408--427. Springer, 2023.

\bibitem[Brank et~al.(2005)Brank, Grobelnik, and Mladenic]{brank2005survey}
Janez Brank, Marko Grobelnik, and Dunja Mladenic.
\newblock A survey of ontology evaluation techniques.
\newblock In \emph{Proceedings of the conference on data mining and data warehouses (SiKDD 2005)}, pages 166--170. Citeseer, 2005.

\bibitem[Buitelaar et~al.(2005)Buitelaar, Cimiano, and Magnini]{buitelaar2005ontology}
Paul Buitelaar, Philipp Cimiano, and Bernardo Magnini.
\newblock \emph{Ontology learning from text: methods, evaluation and applications}, volume 123.
\newblock IOS press, 2005.

\bibitem[Cabot and Navigli(2021)]{cabot2021rebel}
Pere-Llu{\'\i}s~Huguet Cabot and Roberto Navigli.
\newblock Rebel: Relation extraction by end-to-end language generation.
\newblock In \emph{Findings of the Association for Computational Linguistics: EMNLP 2021}, pages 2370--2381, 2021.

\bibitem[Cimiano and V{\"o}lker(2005)]{cimiano2005text2onto}
Philipp Cimiano and Johanna V{\"o}lker.
\newblock Text2onto: A framework for ontology learning and data-driven change discovery.
\newblock In \emph{International conference on application of natural language to information systems}, pages 227--238. Springer, 2005.

\bibitem[Conneau et~al.(2017)Conneau, Kiela, Schwenk, Barrault, and Bordes]{conneau2017supervised}
Alexis Conneau, Douwe Kiela, Holger Schwenk, Lo{\"\i}c Barrault, and Antoine Bordes.
\newblock Supervised learning of universal sentence representations from natural language inference data.
\newblock \emph{arXiv preprint arXiv:1705.02364}, 2017.

\bibitem[Devlin et~al.(2018)Devlin, Chang, Lee, and Toutanova]{devlin2018bert}
Jacob Devlin, Ming-Wei Chang, Kenton Lee, and Kristina Toutanova.
\newblock Bert: Pre-training of deep bidirectional transformers for language understanding.
\newblock \emph{arXiv preprint arXiv:1810.04805}, 2018.

\bibitem[Donahue et~al.(2015)Donahue, Anne~Hendricks, Guadarrama, Rohrbach, Venugopalan, Saenko, and Darrell]{donahue2015long}
Jeffrey Donahue, Lisa Anne~Hendricks, Sergio Guadarrama, Marcus Rohrbach, Subhashini Venugopalan, Kate Saenko, and Trevor Darrell.
\newblock Long-term recurrent convolutional networks for visual recognition and description.
\newblock In \emph{Proceedings of the IEEE conference on computer vision and pattern recognition}, pages 2625--2634, 2015.

\bibitem[Ehrig et~al.(2005)Ehrig, Haase, Hefke, and Stojanovi{\'c}]{Ehrig2005SimilarityFO}
Marc Ehrig, Peter Haase, Mark Hefke, and Nenad Stojanovi{\'c}.
\newblock Similarity for ontologies - a comprehensive framework.
\newblock In \emph{European Conference on Information Systems}, 2005.
\newblock URL \url{https://api.semanticscholar.org/CorpusID:9982461}.

\bibitem[Funk et~al.(2023)Funk, Hosemann, Jung, and Lutz]{funk2023towards}
Maurice Funk, Simon Hosemann, Jean~Christoph Jung, and Carsten Lutz.
\newblock Towards ontology construction with language models.
\newblock \emph{arXiv preprint arXiv:2309.09898}, 2023.

\bibitem[Goyal et~al.(2020)Goyal, Jain, and Ranu]{goyal2020graphgen}
Nikhil Goyal, Harsh~Vardhan Jain, and Sayan Ranu.
\newblock Graphgen: A scalable approach to domain-agnostic labeled graph generation.
\newblock In \emph{Proceedings of The Web Conference 2020}, pages 1253--1263, 2020.

\bibitem[Gruber(1993)]{gruber1993translation}
Thomas~R Gruber.
\newblock A translation approach to portable ontology specifications.
\newblock \emph{Knowledge acquisition}, 5\penalty0 (2):\penalty0 199--220, 1993.

\bibitem[Gruber(1995)]{gruber1995toward}
Thomas~R Gruber.
\newblock Toward principles for the design of ontologies used for knowledge sharing?
\newblock \emph{International journal of human-computer studies}, 43\penalty0 (5-6):\penalty0 907--928, 1995.

\bibitem[Hazman et~al.(2011)Hazman, El-Beltagy, and Rafea]{hazman2011survey}
Maryam Hazman, Samhaa~R El-Beltagy, and Ahmed Rafea.
\newblock A survey of ontology learning approaches.
\newblock \emph{International Journal of Computer Applications}, 22\penalty0 (9):\penalty0 36--43, 2011.

\bibitem[Hearst(1998)]{hearst1998automated}
Marti~A Hearst.
\newblock Automated discovery of wordnet relations.
\newblock \emph{WordNet: an electronic lexical database}, 2, 1998.

\bibitem[Holtzman et~al.(2019)Holtzman, Buys, Du, Forbes, and Choi]{holtzman2019curious}
Ari Holtzman, Jan Buys, Li~Du, Maxwell Forbes, and Yejin Choi.
\newblock The curious case of neural text degeneration.
\newblock \emph{arXiv preprint arXiv:1904.09751}, 2019.

\bibitem[Hu et~al.(2021)Hu, Shen, Wallis, Allen-Zhu, Li, Wang, Wang, and Chen]{hu2021lora}
Edward~J Hu, Yelong Shen, Phillip Wallis, Zeyuan Allen-Zhu, Yuanzhi Li, Shean Wang, Lu~Wang, and Weizhu Chen.
\newblock Lora: Low-rank adaptation of large language models.
\newblock \emph{arXiv preprint arXiv:2106.09685}, 2021.

\bibitem[Jiang et~al.(2023)Jiang, Sablayrolles, Mensch, Bamford, Chaplot, Casas, Bressand, Lengyel, Lample, Saulnier, et~al.]{jiang2023mistral}
Albert~Q Jiang, Alexandre Sablayrolles, Arthur Mensch, Chris Bamford, Devendra~Singh Chaplot, Diego de~las Casas, Florian Bressand, Gianna Lengyel, Guillaume Lample, Lucile Saulnier, et~al.
\newblock Mistral 7b.
\newblock \emph{arXiv preprint arXiv:2310.06825}, 2023.

\bibitem[Karoui et~al.(2004)Karoui, Aufaure, and Bennacer]{karoui2004ontology}
Lobna Karoui, Marie-Aude Aufaure, and Nacera Bennacer.
\newblock Ontology discovery from web pages: Application to tourism.
\newblock In \emph{In the Workshop of Knowledge Discovery and Ontologies}. Citeseer, 2004.

\bibitem[Kashyap et~al.(2005)Kashyap, Ramakrishnan, Thomas, and Sheth]{Kashyap2005TaxaMinerAE}
Vipul Kashyap, Cartic Ramakrishnan, Christopher Thomas, and A.~Sheth.
\newblock Taxaminer: an experimentation framework for automated taxonomy bootstrapping.
\newblock \emph{Int. J. Web Grid Serv.}, 1:\penalty0 240--266, 2005.
\newblock URL \url{https://api.semanticscholar.org/CorpusID:5549251}.

\bibitem[Kaushik and Chatterjee(2018)]{kaushik2018automatic}
Neha Kaushik and Niladri Chatterjee.
\newblock Automatic relationship extraction from agricultural text for ontology construction.
\newblock \emph{Information processing in agriculture}, 5\penalty0 (1):\penalty0 60--73, 2018.

\bibitem[Kingma(2014)]{kingma2014adam}
Diederik~P Kingma.
\newblock Adam: A method for stochastic optimization.
\newblock \emph{arXiv preprint arXiv:1412.6980}, 2014.

\bibitem[Kipf and Welling(2016)]{kipf2016semi}
Thomas~N Kipf and Max Welling.
\newblock Semi-supervised classification with graph convolutional networks.
\newblock \emph{arXiv preprint arXiv:1609.02907}, 2016.

\bibitem[Kuhn(1955)]{kuhn1955hungarian}
Harold~W Kuhn.
\newblock The hungarian method for the assignment problem.
\newblock \emph{Naval research logistics quarterly}, 2\penalty0 (1-2):\penalty0 83--97, 1955.

\bibitem[Kwon et~al.(2023)Kwon, Li, Zhuang, Sheng, Zheng, Yu, Gonzalez, Zhang, and Stoica]{kwon2023efficient}
Woosuk Kwon, Zhuohan Li, Siyuan Zhuang, Ying Sheng, Lianmin Zheng, Cody~Hao Yu, Joseph Gonzalez, Hao Zhang, and Ion Stoica.
\newblock Efficient memory management for large language model serving with pagedattention.
\newblock In \emph{Proceedings of the 29th Symposium on Operating Systems Principles}, pages 611--626, 2023.

\bibitem[Lewis et~al.(2019)Lewis, Liu, Goyal, Ghazvininejad, Mohamed, Levy, Stoyanov, and Zettlemoyer]{lewis2019bart}
Mike Lewis, Yinhan Liu, Naman Goyal, Marjan Ghazvininejad, Abdelrahman Mohamed, Omer Levy, Ves Stoyanov, and Luke Zettlemoyer.
\newblock Bart: Denoising sequence-to-sequence pre-training for natural language generation, translation, and comprehension.
\newblock \emph{arXiv preprint arXiv:1910.13461}, 2019.

\bibitem[Maedche and Staab(2002)]{maedche2002measuring}
Alexander Maedche and Steffen Staab.
\newblock Measuring similarity between ontologies.
\newblock In Asunci{\'o}n G{\'o}mez-P{\'e}rez and V.~Richard Benjamins, editors, \emph{Knowledge Engineering and Knowledge Management: Ontologies and the Semantic Web}, pages 251--263, Berlin, Heidelberg, 2002. Springer Berlin Heidelberg.
\newblock ISBN 978-3-540-45810-4.

\bibitem[Manning et~al.(2014)Manning, Surdeanu, Bauer, Finkel, Bethard, and McClosky]{manning2014stanford}
Christopher~D Manning, Mihai Surdeanu, John Bauer, Jenny~Rose Finkel, Steven Bethard, and David McClosky.
\newblock The stanford corenlp natural language processing toolkit.
\newblock In \emph{Proceedings of 52nd annual meeting of the association for computational linguistics: system demonstrations}, pages 55--60, 2014.

\bibitem[Martello and Toth(1987)]{martello1987linear}
Silvano Martello and Paolo Toth.
\newblock Linear assignment problems.
\newblock In \emph{North-Holland Mathematics Studies}, volume 132, pages 259--282. Elsevier, 1987.

\bibitem[Miller(1995)]{miller1995wordnet}
George~A Miller.
\newblock Wordnet: a lexical database for english.
\newblock \emph{Communications of the ACM}, 38\penalty0 (11):\penalty0 39--41, 1995.

\bibitem[Milo et~al.(2002)Milo, Shen-Orr, Itzkovitz, Kashtan, Chklovskii, and Alon]{milo2002network}
Ron Milo, Shai Shen-Orr, Shalev Itzkovitz, Nadav Kashtan, Dmitri Chklovskii, and Uri Alon.
\newblock Network motifs: simple building blocks of complex networks.
\newblock \emph{Science}, 298\penalty0 (5594):\penalty0 824--827, 2002.

\bibitem[Ponzetto et~al.(2007)Ponzetto, Strube, et~al.]{ponzetto2007deriving}
Simone~Paolo Ponzetto, Michael Strube, et~al.
\newblock Deriving a large scale taxonomy from wikipedia.
\newblock In \emph{AAAI}, volume~7, pages 1440--1445, 2007.

\bibitem[Porzel and Malaka(2004)]{porzel2004task}
Robert Porzel and Rainer Malaka.
\newblock A task-based approach for ontology evaluation.
\newblock In \emph{ECAI Workshop on Ontology Learning and Population, Valencia, Spain}, volume~1. Citeseer Valencia, Spain, 2004.

\bibitem[Raad and Cruz(2015)]{raad2015survey}
Joe Raad and Christophe Cruz.
\newblock A survey on ontology evaluation methods.
\newblock In \emph{Proceedings of the International Conference on Knowledge Engineering and Ontology Development, part of the 7th International Joint Conference on Knowledge Discovery, Knowledge Engineering and Knowledge Management}, 2015.

\bibitem[Reimers and Gurevych(2019)]{reimers-2019-sentence-bert}
Nils Reimers and Iryna Gurevych.
\newblock Sentence-bert: Sentence embeddings using siamese bert-networks.
\newblock In \emph{Proceedings of the 2019 Conference on Empirical Methods in Natural Language Processing}. Association for Computational Linguistics, 11 2019.
\newblock URL \url{http://arxiv.org/abs/1908.10084}.

\bibitem[Roller et~al.(2018)Roller, Kiela, and Nickel]{roller2018hearst}
Stephen Roller, Douwe Kiela, and Maximilian Nickel.
\newblock Hearst patterns revisited: Automatic hypernym detection from large text corpora.
\newblock \emph{arXiv preprint arXiv:1806.03191}, 2018.

\bibitem[Schema.org(2011)]{Schema.org_2011}
Schema.org.
\newblock Schema.org, 2011.
\newblock URL \url{https://www.schema.org/}.

\bibitem[Shen-Orr et~al.(2002)Shen-Orr, Milo, Mangan, and Alon]{shen2002network}
Shai~S Shen-Orr, Ron Milo, Shmoolik Mangan, and Uri Alon.
\newblock Network motifs in the transcriptional regulation network of escherichia coli.
\newblock \emph{Nature genetics}, 31\penalty0 (1):\penalty0 64--68, 2002.

\bibitem[Sorg and Cimiano(2012)]{sorg2012exploiting}
Philipp Sorg and Philipp Cimiano.
\newblock Exploiting wikipedia for cross-lingual and multilingual information retrieval.
\newblock \emph{Data \& Knowledge Engineering}, 74:\penalty0 26--45, 2012.

\bibitem[Sun et~al.(2017)Sun, Ajwani, Nicholson, Sala, and Parthasarathy]{sun2017breaking}
Jiankai Sun, Deepak Ajwani, Patrick~K Nicholson, Alessandra Sala, and Srinivasan Parthasarathy.
\newblock Breaking cycles in noisy hierarchies.
\newblock In \emph{Proceedings of the 2017 ACM on Web Science Conference}, pages 151--160, 2017.

\bibitem[Trajanoska et~al.(2023)Trajanoska, Stojanov, and Trajanov]{trajanoska2023enhancing}
Milena Trajanoska, Riste Stojanov, and Dimitar Trajanov.
\newblock Enhancing knowledge graph construction using large language models.
\newblock \emph{arXiv preprint arXiv:2305.04676}, 2023.

\bibitem[Treeratpituk et~al.(2013)Treeratpituk, Khabsa, and Giles]{Treeratpituk2013GraphbasedAT}
Pucktada Treeratpituk, Madian Khabsa, and C.~Lee Giles.
\newblock Graph-based approach to automatic taxonomy generation (grabtax).
\newblock \emph{ArXiv}, abs/1307.1718, 2013.
\newblock URL \url{https://api.semanticscholar.org/CorpusID:8625171}.

\bibitem[Vercoustre et~al.(2008)Vercoustre, Pehcevski, and Thom]{vercoustre2008using}
Anne-Marie Vercoustre, Jovan Pehcevski, and James~A Thom.
\newblock Using wikipedia categories and links in entity ranking.
\newblock In \emph{Focused Access to XML Documents: 6th International Workshop of the Initiative for the Evaluation of XML Retrieval, INEX 2007 Dagstuhl Castle, Germany, December 17-19, 2007. Selected Papers 6}, pages 321--335. Springer, 2008.

\bibitem[Wang et~al.(2020)Wang, Wei, Dong, Bao, Yang, and Zhou]{wang2020minilm}
Wenhui Wang, Furu Wei, Li~Dong, Hangbo Bao, Nan Yang, and Ming Zhou.
\newblock Minilm: Deep self-attention distillation for task-agnostic compression of pre-trained transformers.
\newblock \emph{Advances in Neural Information Processing Systems}, 33:\penalty0 5776--5788, 2020.

\bibitem[Wei et~al.(2022)Wei, Wang, Schuurmans, Bosma, Xia, Chi, Le, Zhou, et~al.]{wei2022chain}
Jason Wei, Xuezhi Wang, Dale Schuurmans, Maarten Bosma, Fei Xia, Ed~Chi, Quoc~V Le, Denny Zhou, et~al.
\newblock Chain-of-thought prompting elicits reasoning in large language models.
\newblock \emph{Advances in neural information processing systems}, 35:\penalty0 24824--24837, 2022.

\bibitem[Wu et~al.(2019)Wu, Souza, Zhang, Fifty, Yu, and Weinberger]{wu2019simplifying}
Felix Wu, Amauri Souza, Tianyi Zhang, Christopher Fifty, Tao Yu, and Kilian Weinberger.
\newblock Simplifying graph convolutional networks.
\newblock In \emph{International conference on machine learning}, pages 6861--6871. PMLR, 2019.

\bibitem[You et~al.(2018)You, Ying, Ren, Hamilton, and Leskovec]{you2018graphrnn}
Jiaxuan You, Rex Ying, Xiang Ren, William Hamilton, and Jure Leskovec.
\newblock Graphrnn: Generating realistic graphs with deep auto-regressive models.
\newblock In \emph{International conference on machine learning}, pages 5708--5717. PMLR, 2018.

\bibitem[Zavitsanos et~al.(2011)Zavitsanos, Paliouras, and Vouros]{Zavitsanos2011GoldSE}
Elias Zavitsanos, Georgios Paliouras, and George~A. Vouros.
\newblock Gold standard evaluation of ontology learning methods through ontology transformation and alignment.
\newblock \emph{IEEE Transactions on Knowledge and Data Engineering}, 23:\penalty0 1635--1648, 2011.
\newblock URL \url{https://api.semanticscholar.org/CorpusID:15607684}.

\bibitem[Zesch and Gurevych(2007)]{zesch2007analysis}
Torsten Zesch and Iryna Gurevych.
\newblock Analysis of the wikipedia category graph for nlp applications.
\newblock In \emph{Proceedings of the Second Workshop on TextGraphs: Graph-Based Algorithms for Natural Language Processing}, pages 1--8, 2007.

\end{thebibliography}


\appendix
\section{Appendix / supplemental material}

\subsection{Experiment details}  \label{appendix:exp-details}

\subsubsection{Wikipedia}

Some prior works that use Wikipedia categories as a dataset perform additional filtering of concepts considered as ``meta-categories'' mainly used for page management \cite{ponzetto2007deriving}. We instead decided not to further filter the source data to minimise external bias. We note that it is often not clear-cut whether a Wikipedia category is just for page management. For example, the Wikipedia categories of the form ``Lists of [subject]'' refer to the special type of articles where the main body is a bullet point/table listing of the subject, which is a useful concept in the Wikipedia domain.

\subsubsection{\name}  \label{appendix:training-details}

For the Wikipedia experiment, we use Mistral 7B v0.2 (not instruction-tuned) \cite{jiang2023mistral} as the base model. We attach LoRA \cite{hu2021lora} adaptors to all attention and feed-forward layers with parameters $r=32$ and $\alpha=16$. The model is trained for 2 epochs ($\approx$ 17K steps) with batch size 16, context length 2048, and is optimised with Adam using a constant learning rate of 1e-5 with warm-up from zero for the first 100 steps. Finetune uses the same configuration. Training takes 12 A100-hours.

For the arXiv experiment, we further finetune the model trained on Wikipedia with masked loss objective on 2048 document-subgraph pairs from the arXiv training set. We merge the LoRA adaptors from the Wikipedia experiment and initialise new ones with $r=8$ and $\alpha=8$. The model is trained with batch size 16 and Adam with constant learning rate 3e-6 and warp-up from zero for the first 10 steps. Training terminates when the loss stops improving on the evaluation set, which happened at step 288. Finetune (transfer) uses the same configuration. Early stopping happened at step 192.

For both experiments, we finetune the model with the instruction template similar to that of Mistral 7B instruct v0.2. The format is shown below:

\begin{figure}[h]
\centering
\begin{lstlisting}[frame=single]
<s>[INST]\
Title: {{ title }}
{{ abstract }}[/INST]\
{% for path in paths %}
{{ path | join(" -> ") }}
{% endfor %}\
</s>
\end{lstlisting}
\caption{Linearisation template for \name training.}
\label{fig:linearisation-template}
\end{figure}

For inference, we use the vLLM \cite{kwon2023efficient} server which achieves a throughput of $\approx 10$ documents per second. Inference on the validation and test splits of both datasets takes 12 A100-hours in total.

\subsubsection{Hearst}

The Hearst baseline follows the implementation by \citet{roller2018hearst}. Using the tokenization, part-of-speech tagging, lemmatisation, and token regex functionality of the CoreNLP pipeline \cite{manning2014stanford}, taxonomic relations are extracted according to the 28 Hearst patterns used by the authors. Processing all documents takes 10 CPU-hours.

Following the spmi method, low-rank smoothing is applied to the relation matrix to allow comparison between any two concepts even if they are not directly related by an extracted relation. The rank of the smoothed matrix, $r$, is a hyperparameter which we tune by sweeping over $r \in \{5, 10, 15, 20, 25, 50, 100, 150, 200, 250\}$ on the validation set. This defines a dense weighted graph as the raw output. Unfortunately, computing Continuous F1 on a dense graph is very slow, especially for Wikipedia. This is because the Hungarian algorithm used for solving the optimal matching between edges has time complexity $O(N^3)$, where $N$ is the number of edges. To bypass this issue, we perform a pre-filtering step of only exporting the top $10|V|$ weighted edges in the smoothed relation matrix, where $|V|$ is the number of nodes in the graph. For the datasets considered, this density of edges is still much higher than that of the ground truth, and thus, we expect this to have minimal impact on the final output after post-processing.

\subsubsection{REBEL}

We use REBEL-large \cite{cabot2021rebel} in the implementation. The model is an encoder-decoder transformer based on BART-large \cite{lewis2019bart} with 406M parameters. We sample the model with the default configuration used by \citet{cabot2021rebel}. The model is trained to predict 220 types of relations, most of which are not taxonomic relations. We filter the extracted relations and only keep those tagged with ``subclass of'', ``instance of'', ``member of'', and ``part of'' relation types. The same low-rank smoothing method as Hearst is applied to the raw extractions. Processing all documents takes 3 A100-hours.

\subsubsection{Prompting}

To obtain more comparable results, we use Mistral 7B Instruct v0.2, the instruction-tuned version of the base model of \name, as the LLM for our prompting baseline. For One-shot and Three-shot, we randomly sample examples from the training set for each query. The output is parsed using regex and results that do not match the regex are discarded. We perform manual prompt engineering by inspecting individual responses. The final prompt template is shown in \cref{fig:prompt-template}. The total inference cost for all prompting baselines is $\approx 50$ A100-hours.

\begin{figure}
\centering
\begin{lstlisting}[frame=single]
The following is an article's title and abstract. Your task is to assign this article to suitable category hierarchy. A category is typically represented by a word or a short phrase, representing broader topics/concepts that the article is about. A category hierarchy represented by a collection of paths from the generic root category "Main topic classifications" to a specific category suitable for the article. The topics titles should become more and more specific as you move from the root to the leaf. 

{% if examples|length > 0 %}
{% for example in examples %}
### EXAMPLE {{ loop.index }} ###
### ARTICLE ###
Title: {{ example['title'] }}
{{ example['abstract'] }}
### END ARTICLE ###
{% for path in example['paths'] %}
{{ path | join(" -> ") }}
{% endfor %}
### END EXAMPLE {{ loop.index }} ###
{% endfor %}
{% else %}
You must answer in the format of:
Main topic classifications -> Broad topic 1 -> Subtopic 1 -> ... -> Most specific topic 1
Main topic classifications -> Borad topic 2 -> Subtopic 2 -> ... -> Most specific topic 2
...
{% endif %}

### ARTICLE ###
Title: {{ title }}
{{ abstract }}
### END ARTICLE ###

Provide a category hierarchy for the above article. \
{% if examples|length > 0 %}
Use the same format as the examples above.
{% else %}
Use the format described above.
{% endif %}
\end{lstlisting}
\caption{Prompt template used for the Zero/One/Three-shot baselines.}
\label{fig:prompt-template}
\end{figure}

\newpage
\subsubsection{Hyperparameters}

The raw generated outputs of all methods are post-processed with the same scheme as described in \cref{sec:method:post-processing}. The best hyperparameters for the post-processing step found by grid search on the validation are reported in \cref{tab:hyperparams}.

\begin{table}[t]
\centering
\captionsetup{width=.9\linewidth}
\caption{Values of the best hyperparameters found by grid search. $r$ is the rank of the low-rank smoothing, only applicable to Hearst and REBEL. $\alpha = \beta = 0$ means no edges are pruned from the raw output apart from self-loop and inverse edge removal.}
\label{tab:hyperparams}
\begin{tabular}{lllll}
    \toprule
    Dataset & Method & $\alpha$ & $\beta$ & $r$ \\
    \midrule
    \multirow[t]{8}{*}{Wikipedia}
     & Memorisation & 0 & 0.058489 & - \\
     & Hearst & 0.786685 & 0 & 5 \\
     & REBEL & 0.872544 & 0 & 20 \\
     & Zero-shot & 0.976781 & 0.298107 & - \\
     & One-shot & 0.990906 & 0.346684 & - \\
     & Three-shot & 0.991955 & 0.530957 & - \\
     & Finetune & 0.883848 & 0.058489 & - \\
     & \name & 0.974330 & 0.025893 & - \\
    \midrule
    \multirow[t]{8}{*}{arXiv}
     & Memorisation & 0.340246 & 0 & - \\
     & Hearst & 0.595878 & 0 & 150 \\
     & REBEL & 0.836685 & 0 & 100 \\
     & Zero-shot & 0.999896 & 0.346684 & - \\
     & One-shot & 0.999611 & 0.401187 & - \\
     & Three-shot & 0.999851 & 0.298107 & -\\
     & Finetune (transfer) & 0.988129 & 0.346684 & - \\
     & \name (transfer) & 0.983681 & 0.123872 & - \\
    \bottomrule
\end{tabular}
\end{table}

\newpage
\subsection{Ablations}

In this section, we present the results of our ablations regarding output consistency, the benefits of more advanced prompting techniques, and a comparison against LLMs4OL \cite{babaei2023llms4ol}.

\subsubsection{Consistency}
\label{appendix:consistency}

A common assumption of taxonomic relations is its transitivity and anti-symmetry. One limitation of many OL methods, including OLLM, is that they do not guarantee that the generated ontology is cycle-free, leading to inconsistent taxonomic relations. To achieve consistency, generic post-processing techniques \cite{sun2017breaking} can be applied to remove such cycles.

We analysed the ontologies generated by OLLM and found only 97 simple cycles in Wikipedia and none in arXiv. Using the greedy algorithm of repeatedly removing the edge that breaks the most simple cycles (a heuristic to the smallest set of edges whose removal makes the graph acyclic), we prune all such cycles and make the ontology consistent by removing just 26 of 10414 edges in Wikipedia. This is surprising considering we did not explicitly optimise our model to satisfy consistency.

\subsubsection{Chain of thought prompting}

More sophisticated prompting techniques, such as chain-of-thought (CoT) \cite{wei2022chain} have been shown to bring significant improvements in LLM inference. We explore whether we can establish strong baselines here by employing CoT in our prompting methods.

We extend the zero-shot prompting method such that prediction now involves two rounds of inference: In the first round, we ask the model to describe the possible relevant concepts for the input document and to explain its reasoning. Then, we ask the model to predict the subgraph in the specified format given the additional, self-generated context. The prompts used are shown below:

\begin{figure}[h]
\centering
\begin{lstlisting}[frame=single]
The following is an article's title and abstract. Briefly break down the topics (both specific and general concepts) relevant to this article. Explain your reasoning step by step.

### ARTICLE ###
Title: {{ title }}
{{ abstract }}
### END ARTICLE ###
\end{lstlisting}
\caption{Chain-of-thought first prompt}
\end{figure}

\begin{figure}[h]
\centering
\begin{lstlisting}[frame=single]
Your task now is to assign this article to a suitable category hierarchy. A category is typically represented by a word or a short phrase, representing broader topics/concepts that the article is about. A category hierarchy is represented by a collection of paths from the generic root category "Main topic classifications" to a specific category suitable for the article. The topic titles should become more and more specific as you move from the root to the leaf.

You must answer in the format of:
Main topic classifications -> Broad topic 1 -> Subtopic 1 -> ... -> Most specific topic 1
Main topic classifications -> Broad topic 2 -> Subtopic 2 -> ... -> Most specific topic 2
...
\end{lstlisting}
\caption{Chain-of-thought second prompt}
\end{figure}

We tested the CoT method on Wikipedia and found no significant difference from basic zero-shot prompting, as shown in \cref{table:cot-metrics}. We attribute this to the fact that CoT prompting primarily aims to improve logic and reasoning. We hypothesise that the performance in OL is more dependent on the model’s understanding of natural language than its ability to perform multi-step reasoning, hence we do not observe any significant improvement from CoT.

\addtolength{\tabcolsep}{-0.2em}
\begin{table}[t]
\caption{Comparison of Zero-shot with and without chain-of-thought prompting. There is no significant difference in performance.}
\label{table:cot-metrics}
\centering
\begin{small}
\begin{tabularx}{\linewidth}{l l X X X X l}
\toprule
Dataset & Method & Literal F1 $\uparrow$ & Fuzzy F1 $\uparrow$ & Cont. F1 $\uparrow$ & Graph F1 $\uparrow$ & Motif Dist. $\downarrow$ \\
\midrule
\multirow[t]{8}{*}{Wikipedia} & 
   Zero-shot & \textbf{0.007} & 0.871 & \textbf{0.455} & \textbf{0.639} & \textbf{0.341} \\
 & Zero-shot CoT & \textbf{0.007} & \textbf{0.873} & 0.449 & 0.635 & 0.357 \\
\bottomrule
\end{tabularx}
\end{small}
\end{table}

\subsubsection{Comparison against LLMs4OL}
\label{appendix:llms4ol}

In this ablation, we evaluate whether the improvement by OLLM is due to the improved methodology (end-to-end modelling) or simply due to the use of LLMs. One way to construct ontologies with LLMs proposed by LLMs4OL is to first prompt LLMs for possible concepts in a document, then link prediction by prompting for a yes/no response. Unfortunately, constructing a baseline from such two subtasks is non-trivial. We encountered significant scalability issues in the link prediction stage as it required $O(n^2)$ inferences. We make two modifications to overcome such limitation:
\begin{enumerate}[itemsep=0pt,leftmargin=*]
    \item After the concept discovery stage, we only discard all but the $n$ most frequent concepts to limit the number of inferences required during link prediction, where $n$ is the number of concepts in the ground truth.
    \item Instead of using zero-shot Mistral 7B as the link predictor, we use a finetuned BERT as the link predictor as it runs much faster. Given that LLMs4OL demonstrated that finetuned models perform much better than zero-shot inference on link prediction, we expect the finetuned BERT to be at least as good, if not better, than zero-shot Mistral 7B on this subtask.
\end{enumerate}

We design this ablation such that it is comparable to zero-shot end-to-end modelling: both use zero-shot Mistral 7B as the backbone, just utilised in different ways. We tested this method on Wikipedia and found that it is worse than zero-shot end-to-end modelling on all metrics except Motif Distance, as shown in \cref{table:llms4ol}. This is evidence that our end-to-end modelling approach is a clear improvement over traditional subtask-based OL. Not only does LLMs4OL suffer from significant scalability bottlenecks thus unlikely to be scalable to solve large problems, its performance is also worse. The results suggest that we can more effectively and efficiently leverage the capabilities of LLMs beyond just solving subtasks, such as by predicting subgraphs.

\addtolength{\tabcolsep}{-0.2em}
\begin{table}[t]
\caption{Comparison of Zero-shot end-to-end modelling and LLMs4OL-style modelling with zero-shot concept discovery and fine-tuned BERT link prediction. LLMs4OL generally performs worse than zero-shot.}
\label{table:llms4ol}
\centering
\begin{small}
\begin{tabularx}{\linewidth}{l l X X X X l}
\toprule
Dataset & Method & Literal F1 $\uparrow$ & Fuzzy F1 $\uparrow$ & Cont. F1 $\uparrow$ & Graph F1 $\uparrow$ & Motif Dist. $\downarrow$ \\
\midrule
\multirow[t]{8}{*}{Wikipedia} & 
   Zero-shot & \textbf{0.007} & \textbf{0.871} & \textbf{0.455} & \textbf{0.639} & 0.341 \\
 & LLMs4OL & 0.003 & 0.841 & 0.428 & 0.482 & \textbf{0.092} \\
\bottomrule
\end{tabularx}
\end{small}
\end{table}

\newpage
\subsection{Visualising evaluation metrics}

\subsubsection{Visualisation of node matching in Graph F1}

\begin{figure}[h]
    \centering
    \includegraphics[width=\textwidth]{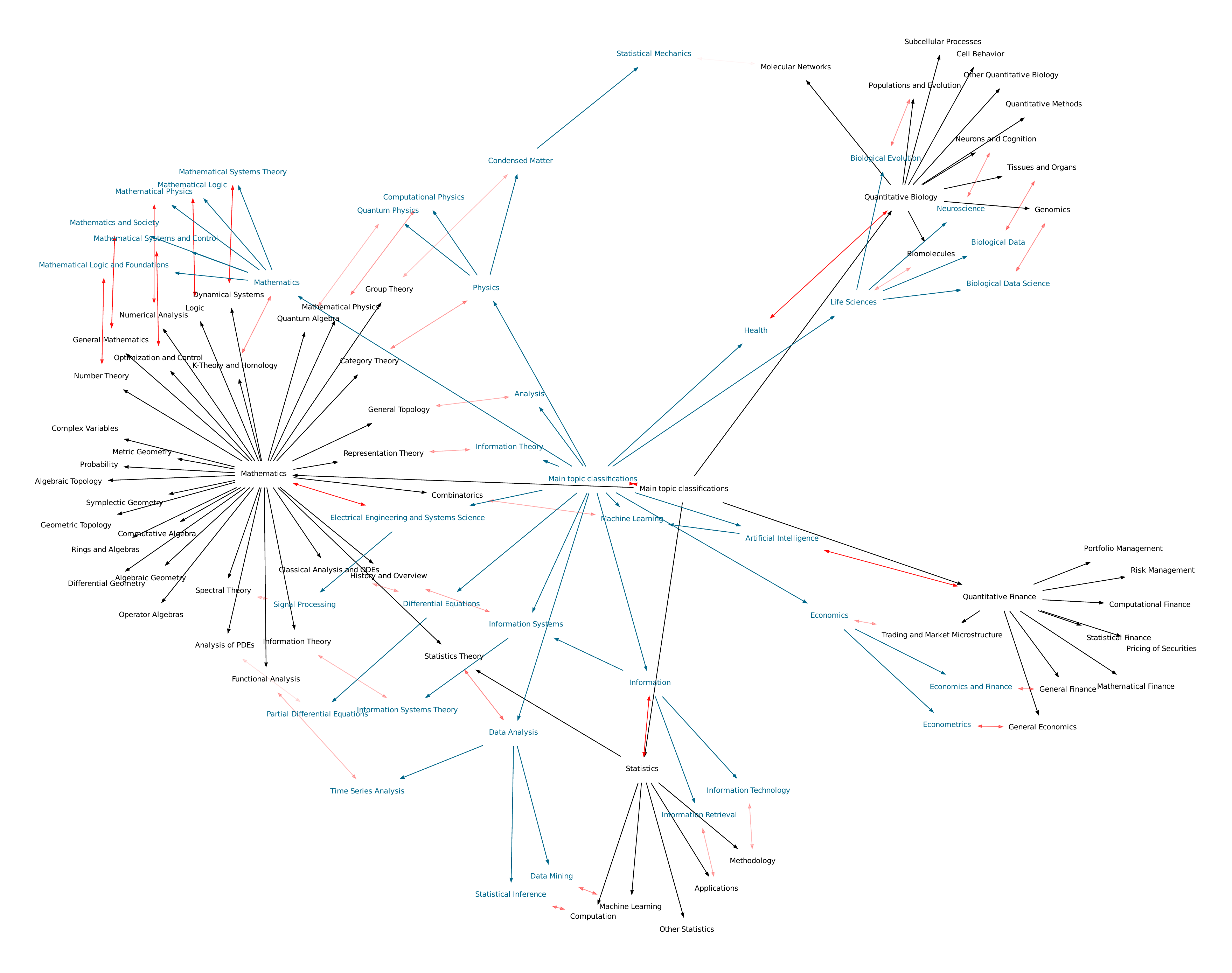}
    \caption{Highest scoring node matching from the Graph F1 metric between the ontology generated by \name (\textcolor{teal}{\textbf{teal}}) and the ground truth ontology (\textbf{black}). The matching between nodes is shown in \textcolor{red}{\textbf{red}}, where the opacity of the edge indicates the similarity score (weaker links are more transparent). Visually, the matching defines a clear alignment of the two graphs: from the centre to the left we have the Mathematics-related concepts; at the top right we have Biology-related concepts; and at the bottom right we have Economics-related concepts.}
    \label{fig:graph-matching}
\end{figure}

\newpage
\subsubsection{Visualisation of edge matching in Continuous F1}

\begin{figure}[h]
    \centering
    \includegraphics[width=\textwidth]{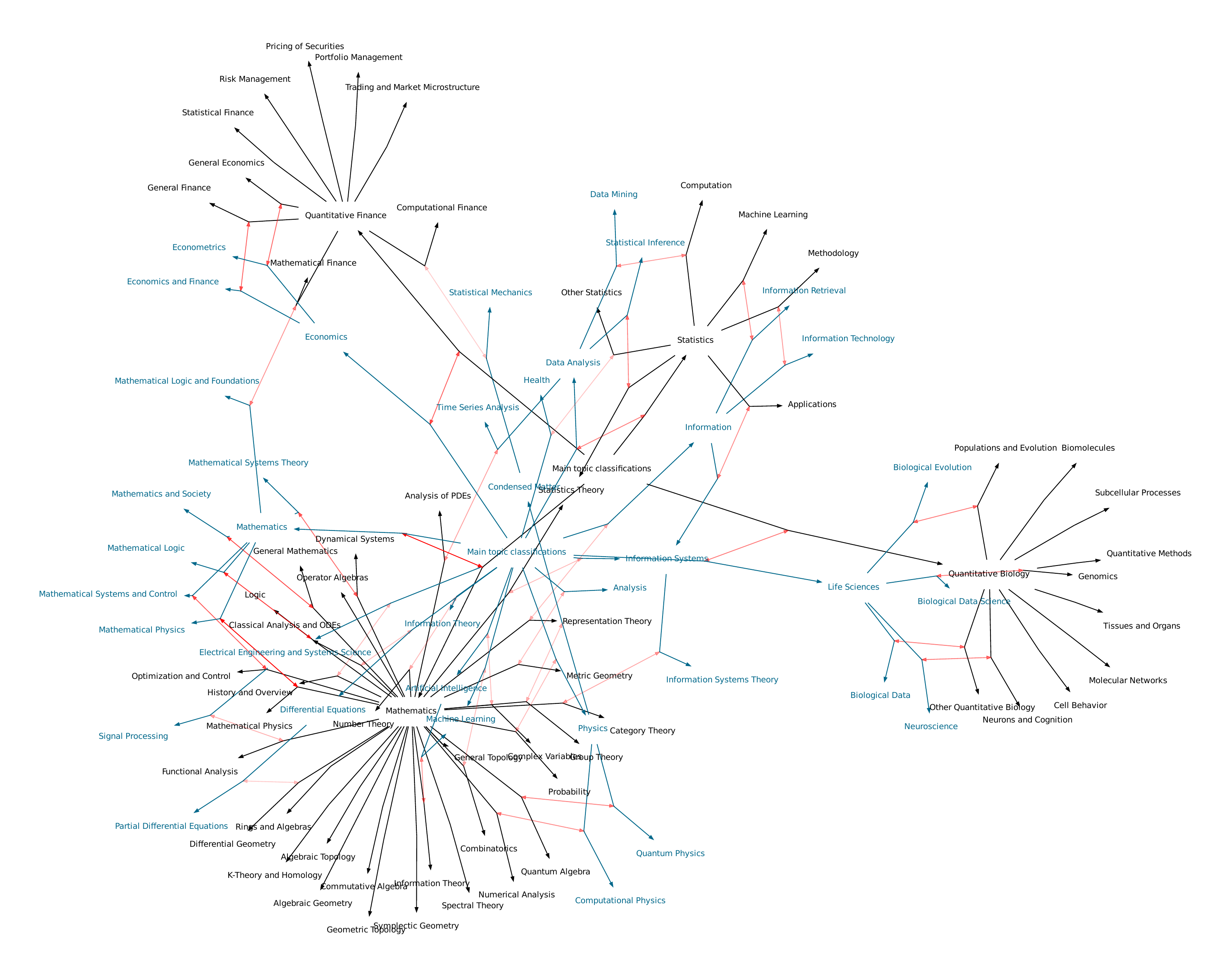}
    \caption{Highest scoring edge matching from the Continuous F1 metric between the ontology generated by \name (\textcolor{teal}{\textbf{teal}}) and the ground truth ontology (\textbf{black}). The matching between edges is shown in \textcolor{red}{\textbf{red}}, where the opacity of the edge indicates the similarity score (weaker links are more transparent). Visually, the matching defines a clear alignment of the two graphs: in the bottom left and centre we have the Mathematics-related concepts; at the right we have Biology-related concepts; and at the top left we have Economics-related concepts.}
    \label{fig:edge-matching}
\end{figure}

\subsection{Visualisation of generated ontologies} \label{appendix:visualisation}

\subsubsection{Wikipedia} \label{appendix:viz-wiki}

We include some generated outputs for Wikipedia here. Since the full generated output is too large to visualise, we plot subgraphs of the output instead. We sample the subgraphs by the following method:
\begin{enumerate}
    \item Pick a random node in the generated graph. 
    \item Get the induced subgraph by the 1-hop neighbourhood of the chosen node.
    \item Include the shortest path from the root ``Main topic classifications'' to the chosen node if such path exists.
    \item Repeat from step 1 if the subgraph has more than 30 nodes or less than 5 nodes.
\end{enumerate}
We apply the filtering step (step 4) as subgraphs with too many nodes are difficult to inspect manually, and those with too few are uninformative. For Hearst, we choose the filtering upper bound to be 50 nodes as we fail to find subgraphs smaller than 30 nodes quickly. We additionally colour each edge \textcolor{black}{\textbf{black}} if it occurs literally in the training graph, \textcolor{blue}{blue} if it occurs literally in the test graph, and \textcolor{red}{red} otherwise.

\newpage
\begin{figure}[H]
    \centering
    \begin{subfigure}{0.9\textwidth}
    \centering
    \includegraphics[width=\linewidth]{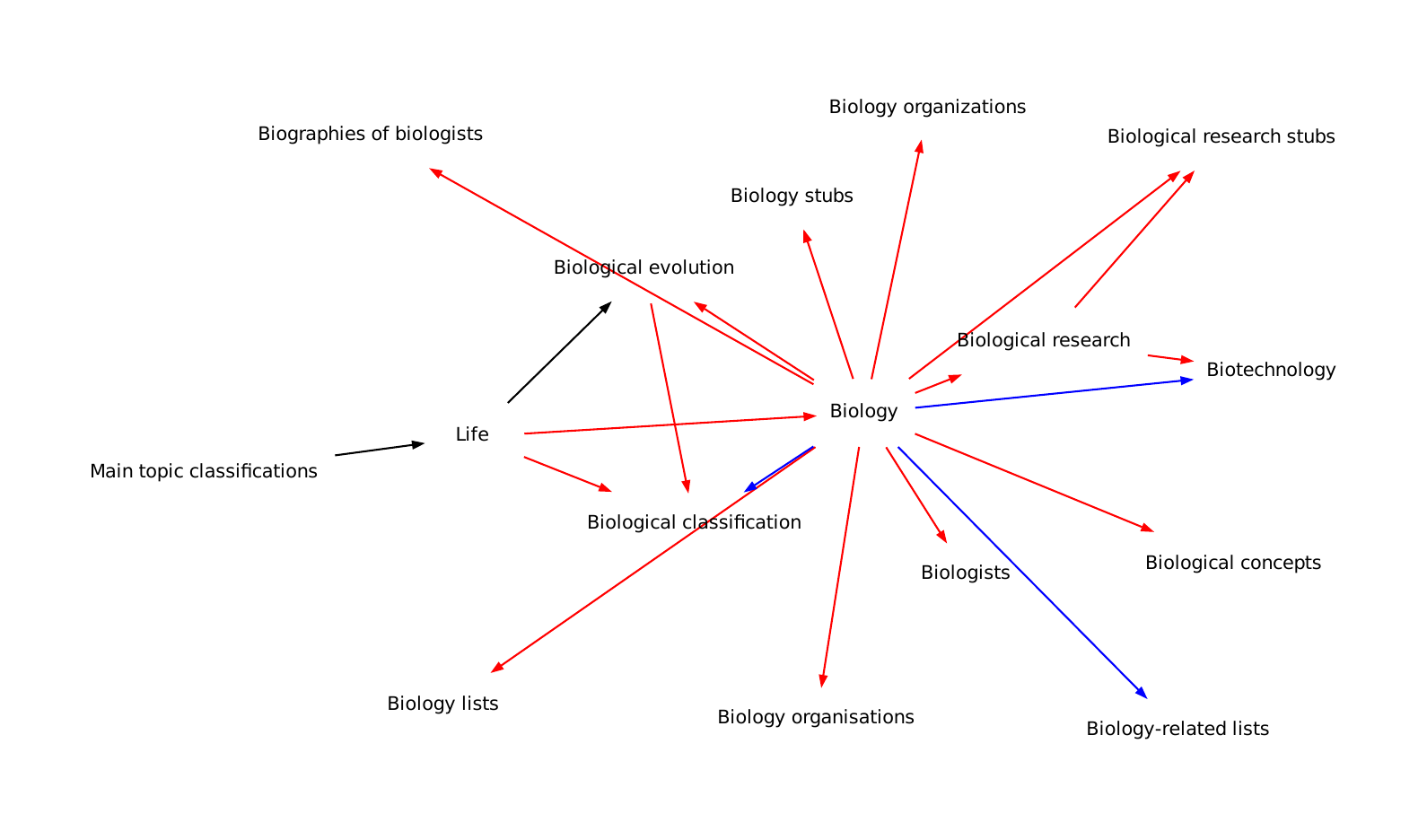}
    \caption{Biology}
    \end{subfigure}
    \begin{subfigure}{0.9\textwidth}
    \centering
    \includegraphics[width=\linewidth]{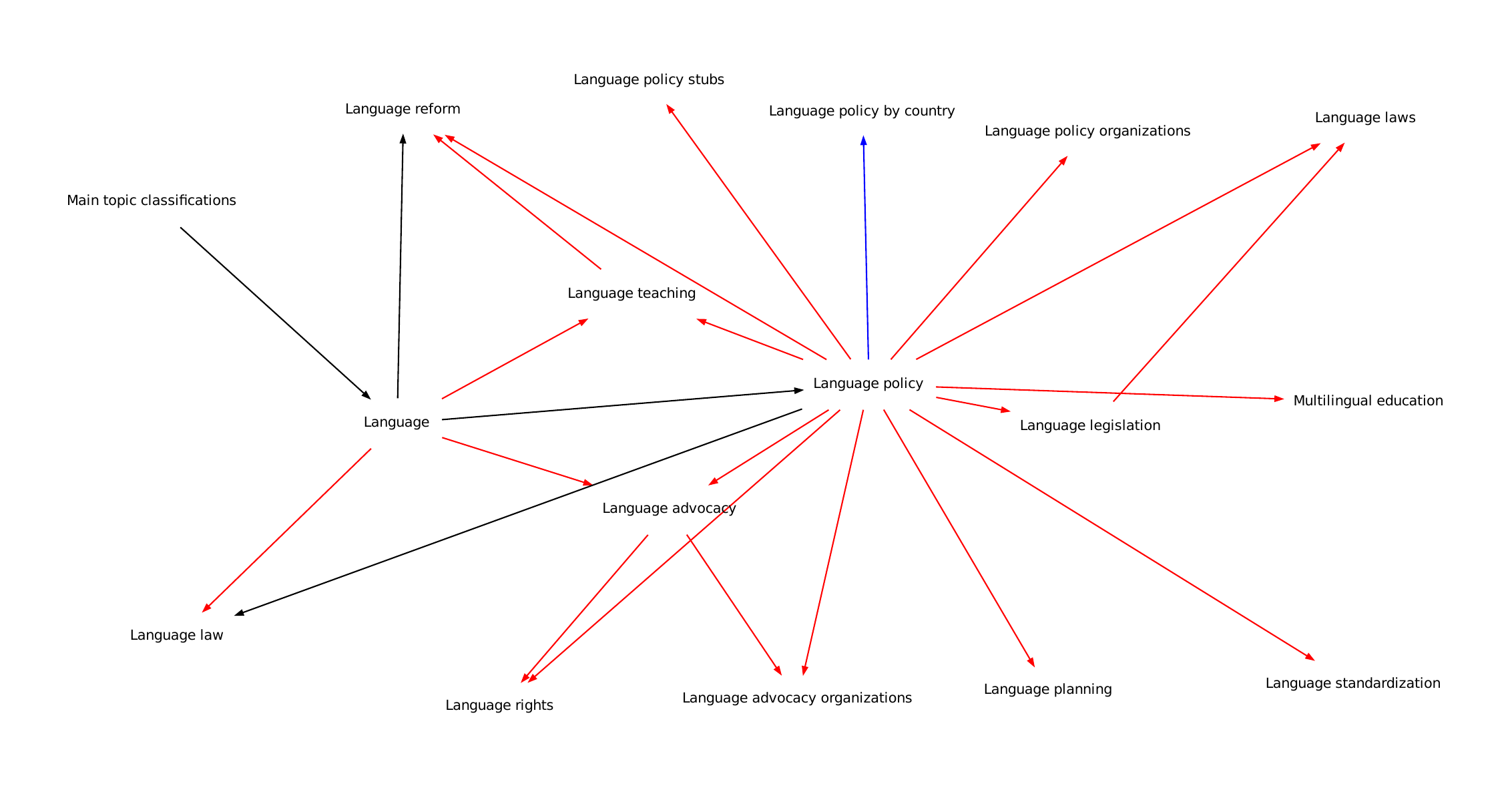}
    \caption{Language policy}
    \end{subfigure}
    \begin{subfigure}{0.9\textwidth}
    \centering
    \includegraphics[width=\linewidth]{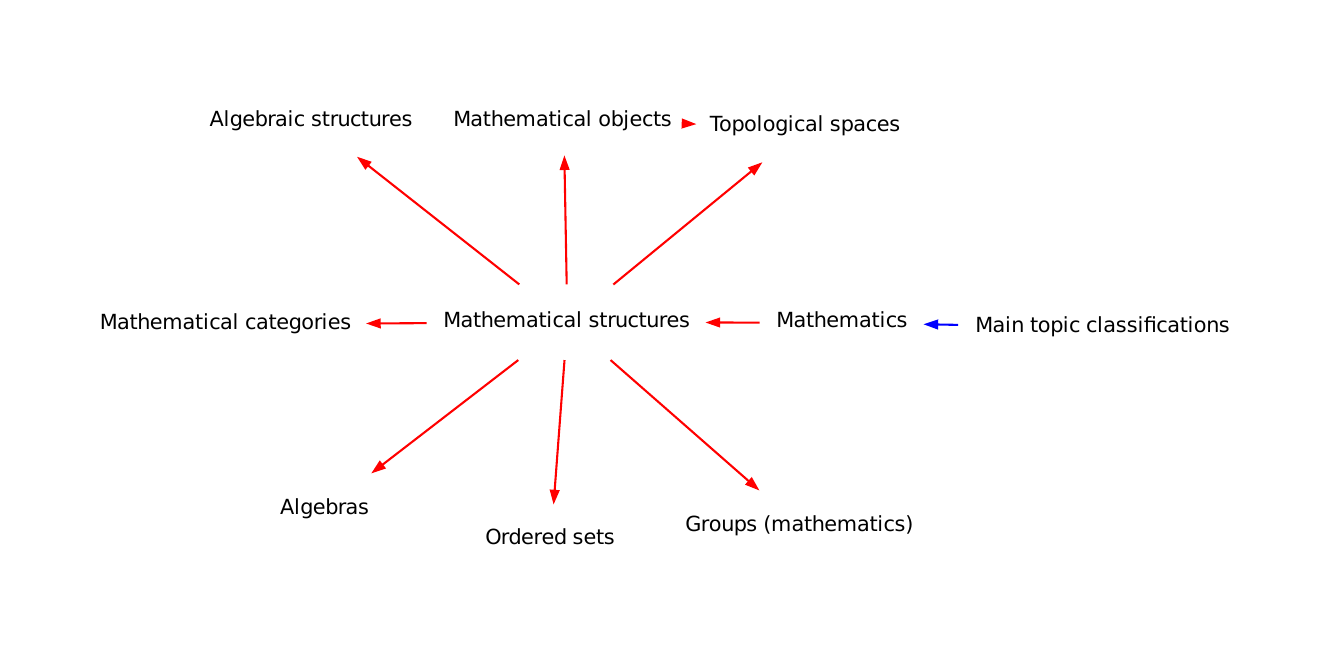}
    \caption{Mathematical structures}
    \label{fig:ollm-wiki-samples-math}
    \end{subfigure}
    \caption{Sub-ontologies for Wikipedia generated by \name, centred on various topics.}
\end{figure}

\begin{figure}[H]
    \centering
    \begin{subfigure}{1.0\textwidth}
    \centering
    \includegraphics[width=\linewidth]{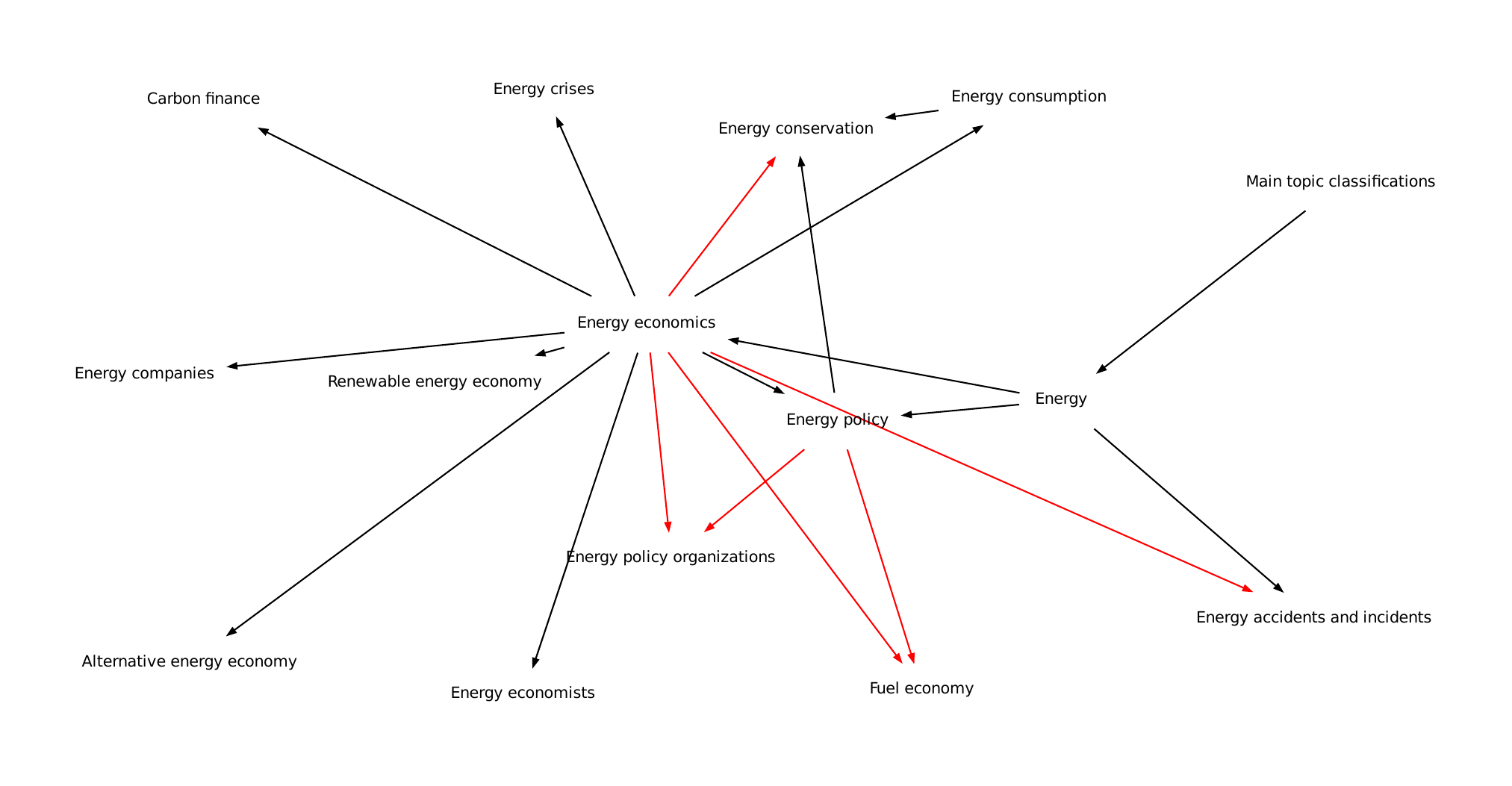}
    \caption{Energy economics}
    \end{subfigure}
    \begin{subfigure}{1.0\textwidth}
    \centering
    \includegraphics[width=\linewidth]{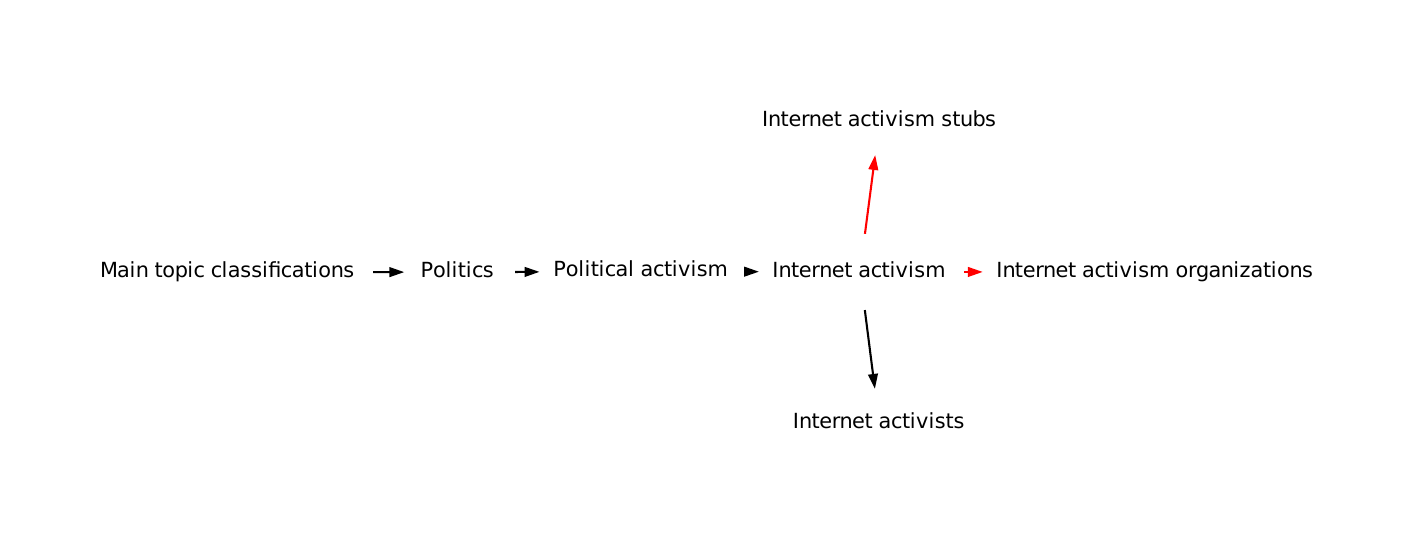}
    \caption{Internet activism}
    \end{subfigure}
    \begin{subfigure}{1.0\textwidth}
    \centering
    \includegraphics[width=\linewidth]{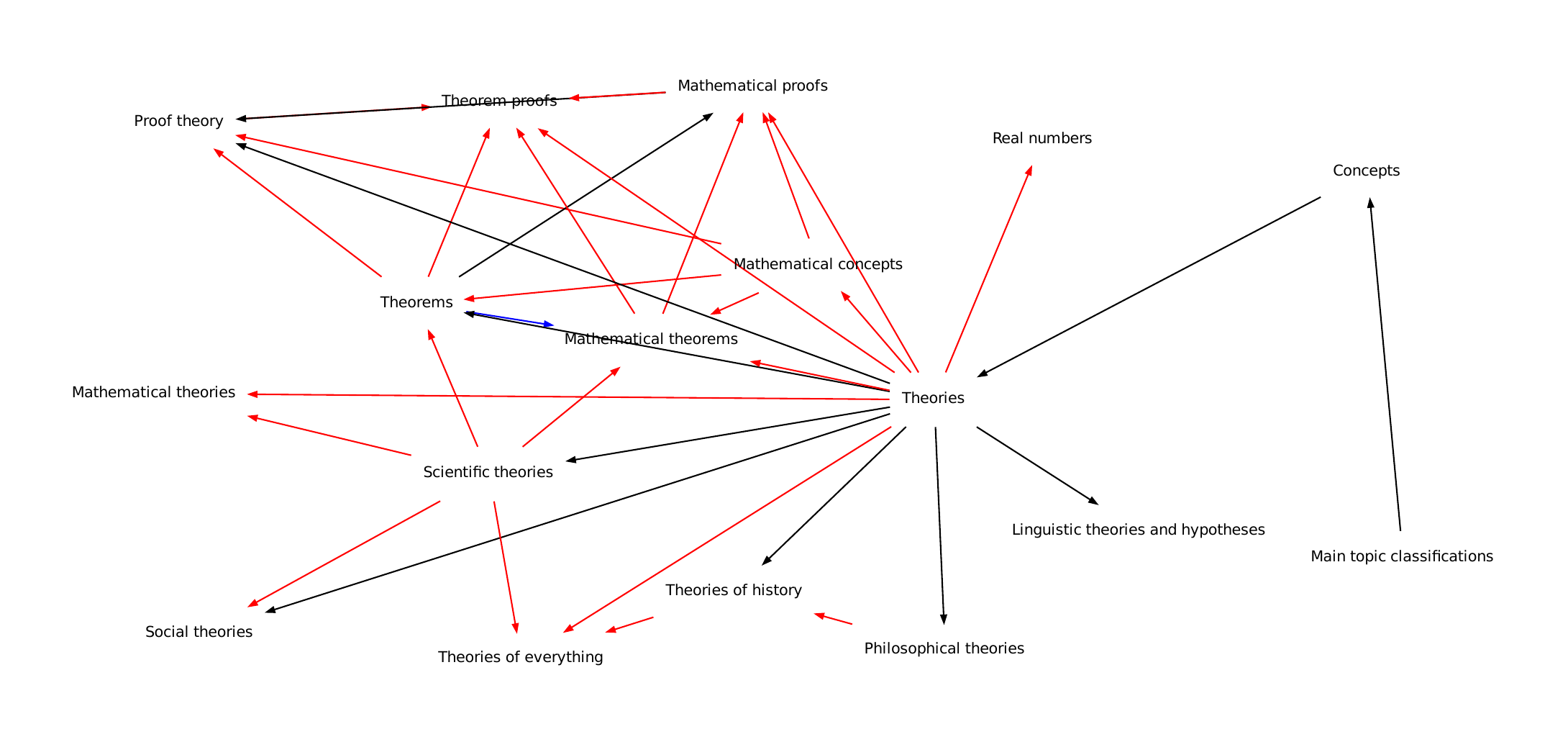}
    \caption{Theories}
    \end{subfigure}
    \caption{Sub-ontologies for Wikipedia generated by Finetune, centred on various topics.}
\end{figure}

\begin{figure}[H]
    \centering
    \begin{subfigure}{1.0\textwidth}
    \centering
    \includegraphics[width=\linewidth]{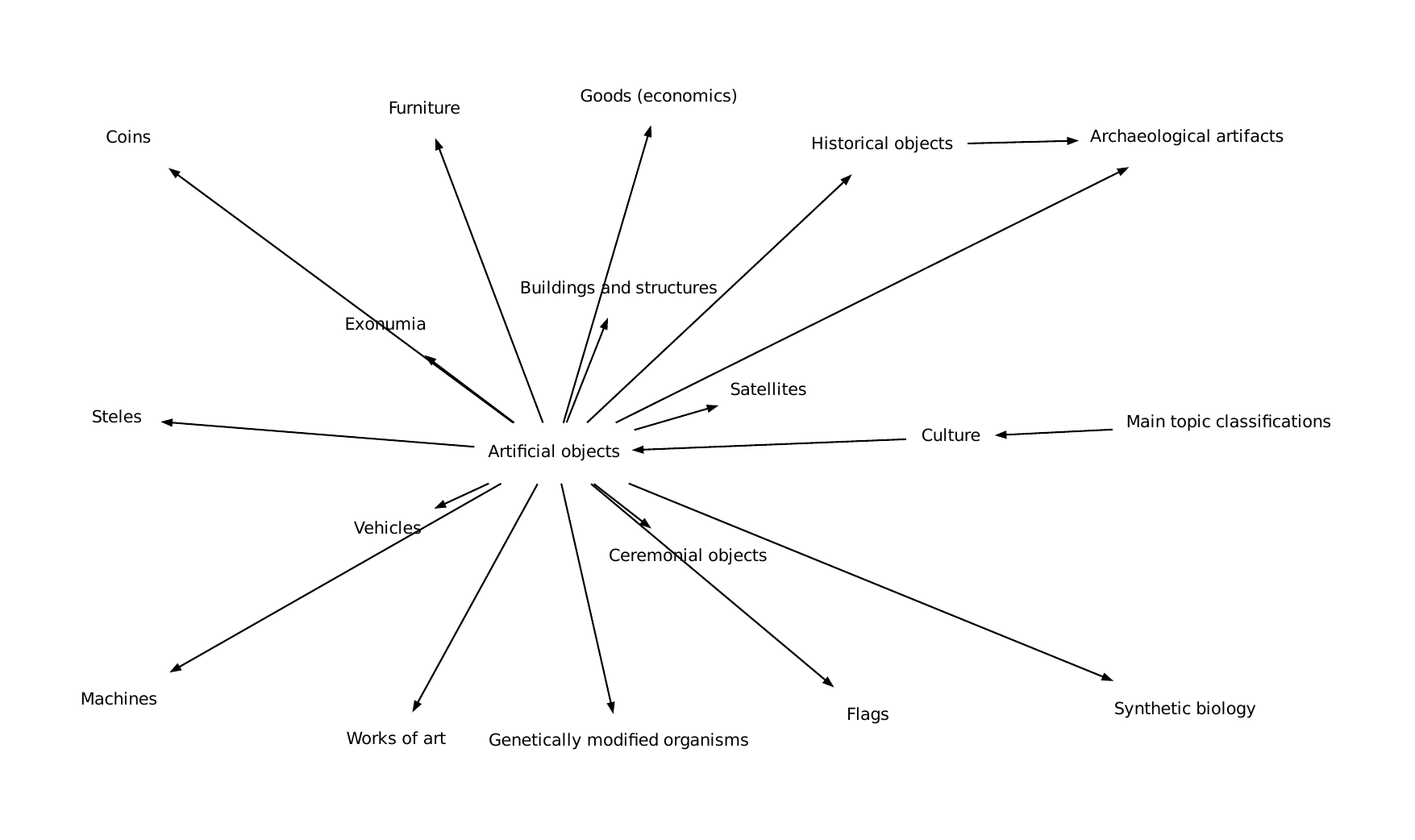}
    \caption{Aritificial objects}
    \end{subfigure}
    \begin{subfigure}{0.9\textwidth}
    \centering
    \includegraphics[width=\linewidth]{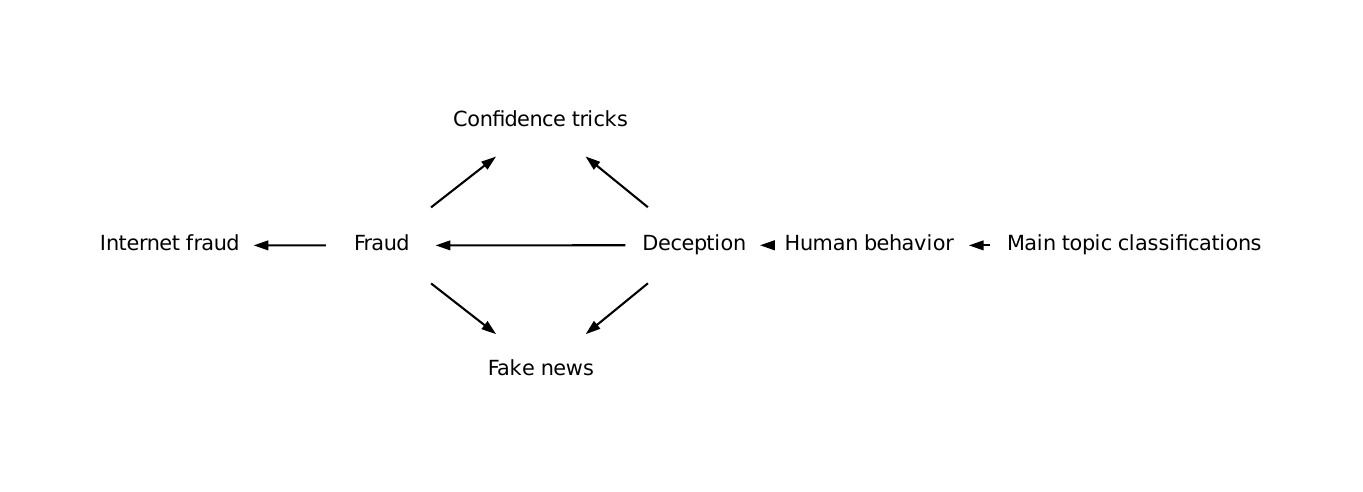}
    \caption{Fraud}
    \end{subfigure}
    \begin{subfigure}{0.9\textwidth}
    \centering
    \includegraphics[width=\linewidth]{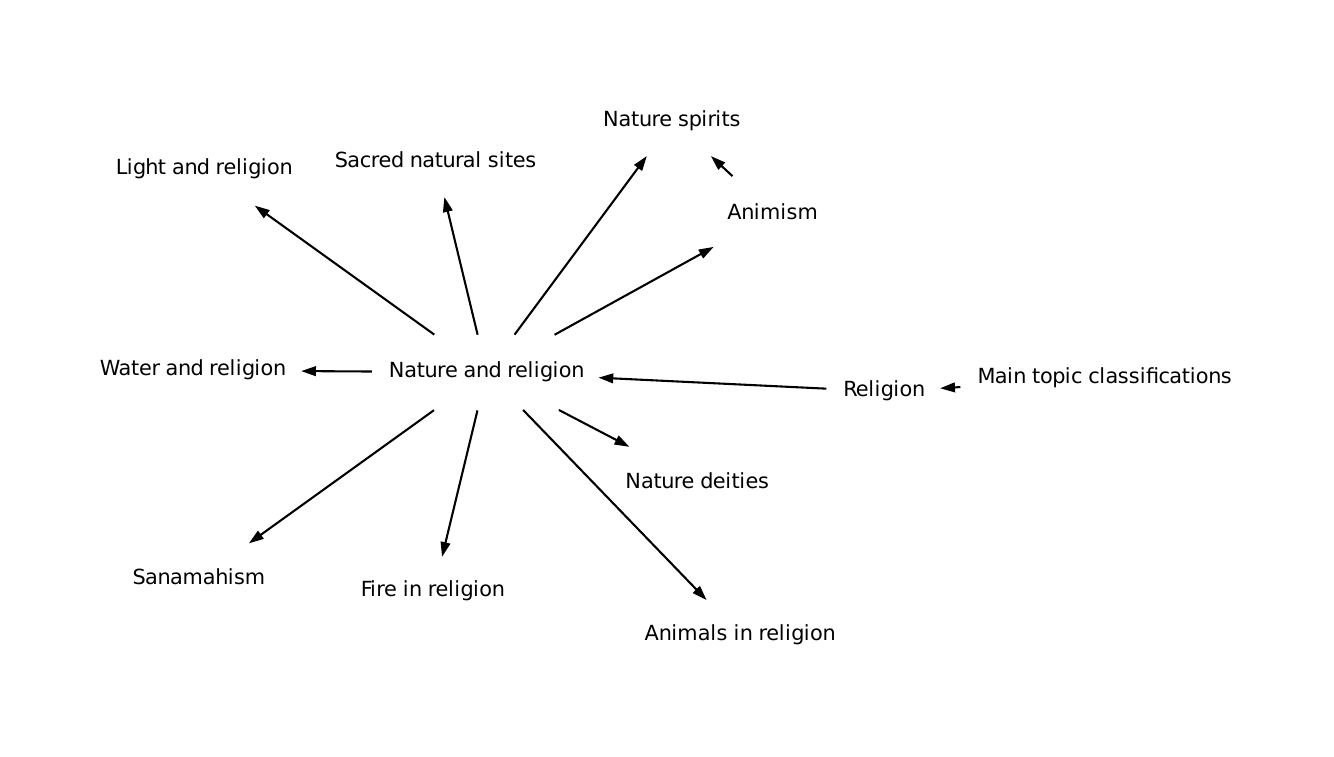}
    \caption{Nature and religion}
    \end{subfigure}
    \caption{Sub-ontologies for Wikipedia generated by Memorisation, centred on various topics.}
\end{figure}

\begin{figure}[H]
    \centering
    \begin{subfigure}{0.7\textwidth}
    \centering
    \includegraphics[width=\linewidth]{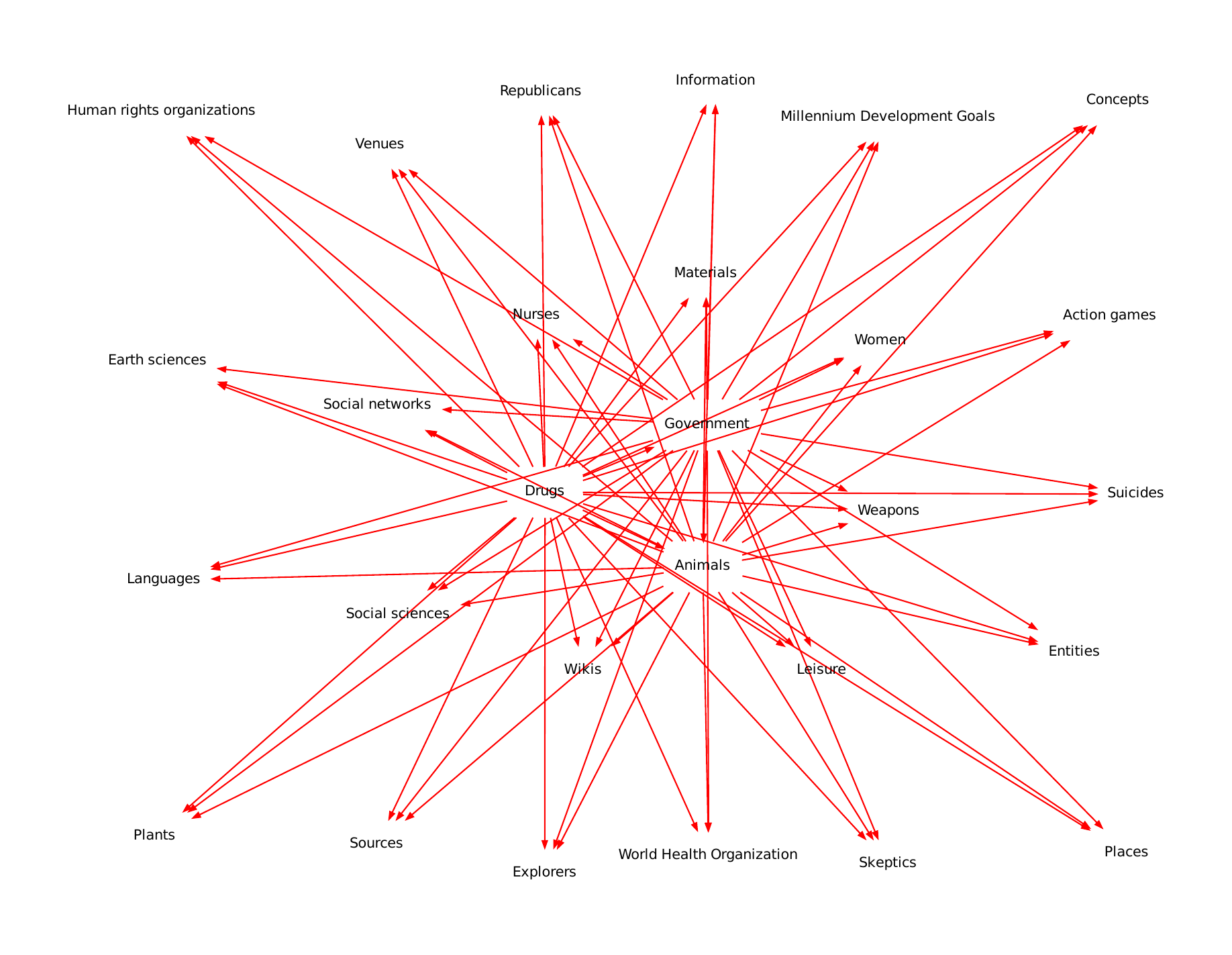}
    \caption{Drugs}
    \end{subfigure}
    \begin{subfigure}{0.6\textwidth}
    \centering
    \includegraphics[width=\linewidth]{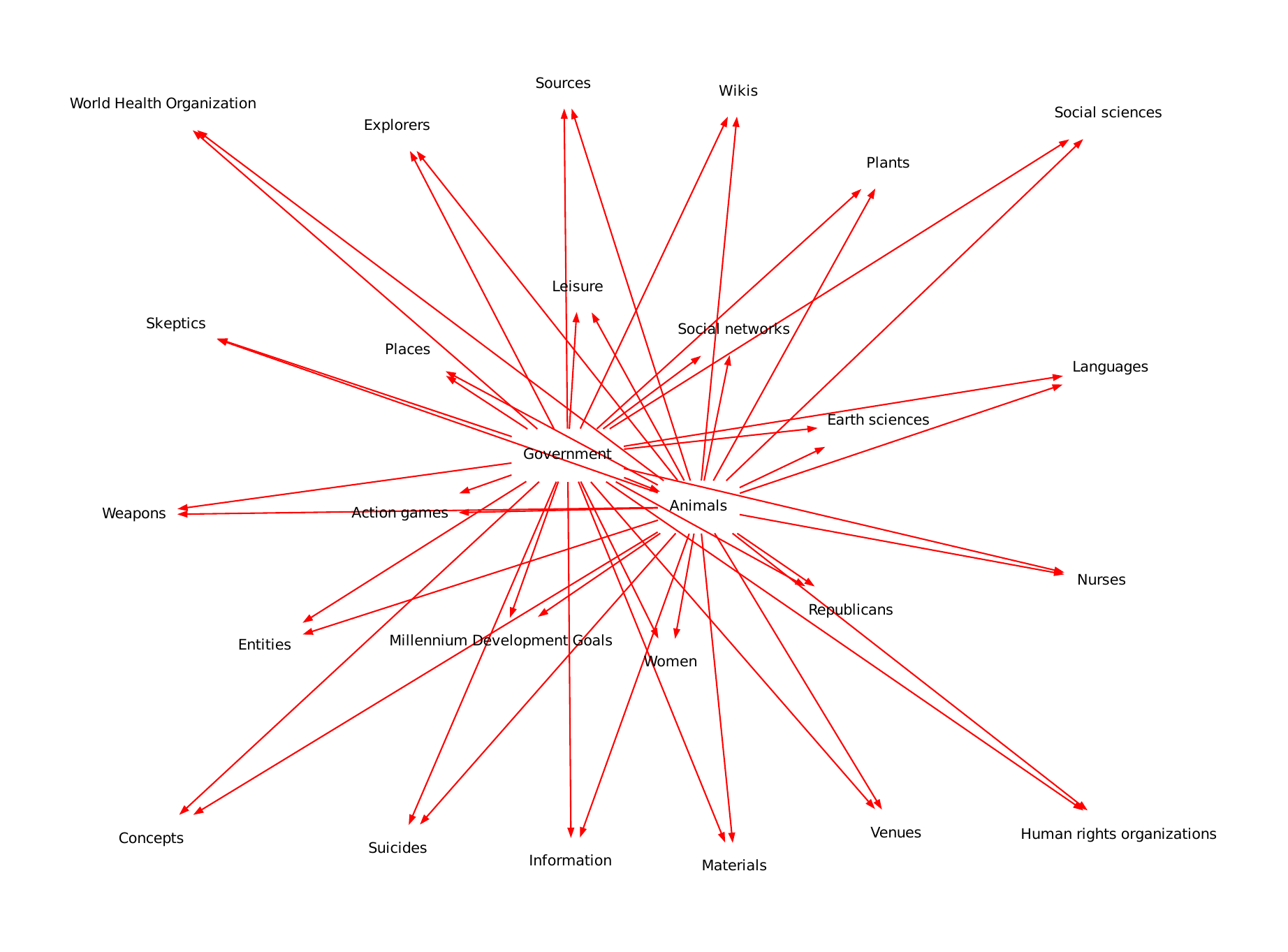}
    \caption{Government}
    \end{subfigure}
    \begin{subfigure}{0.7\textwidth}
    \centering
    \includegraphics[width=\linewidth]{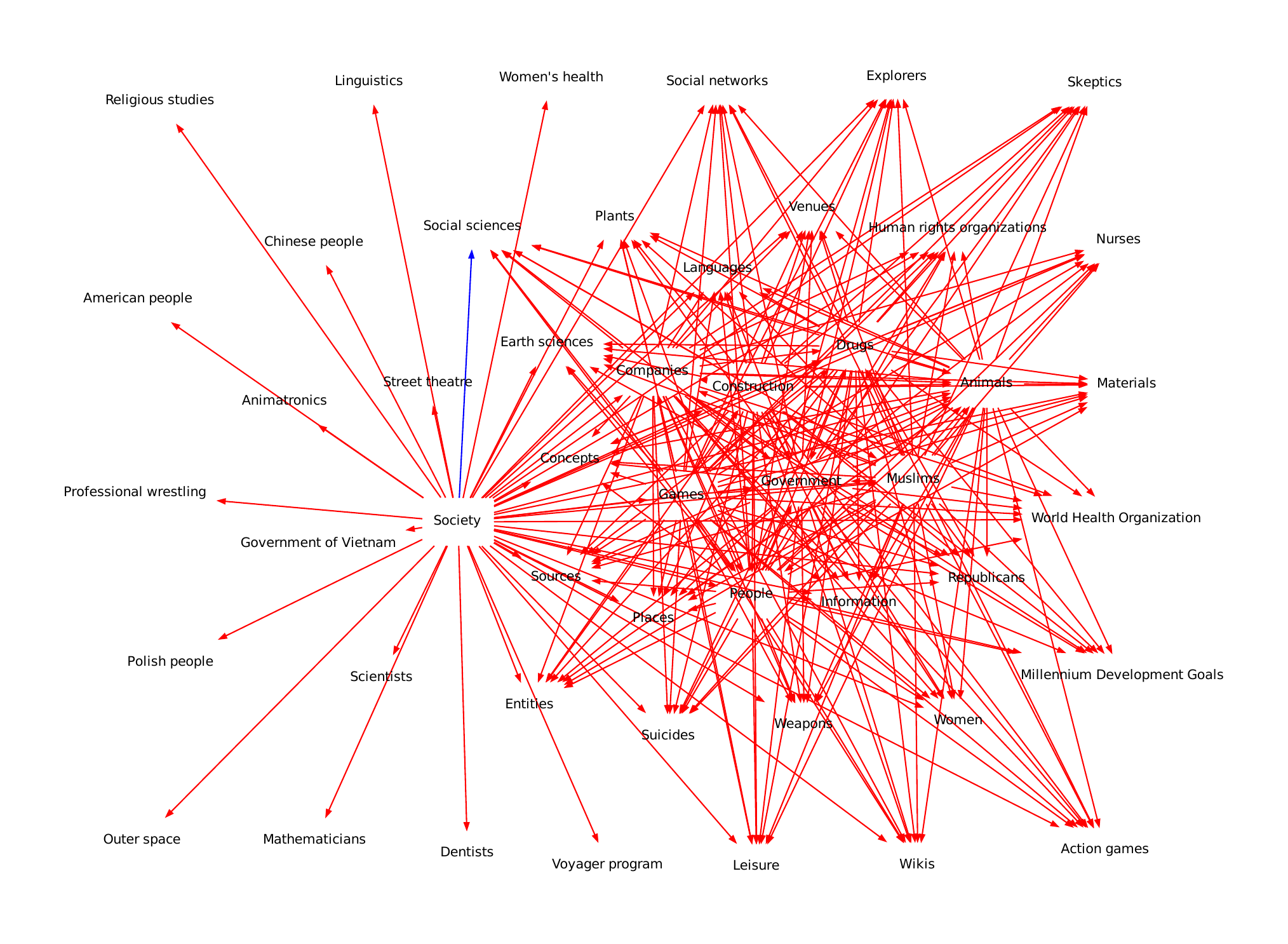}
    \caption{Society}
    \end{subfigure}
    \caption{Sub-ontologies for Wikipedia generated by Hearst, centred on various topics.}
\end{figure}

\begin{figure}[H]
    \centering
    \begin{subfigure}{0.9\textwidth}
    \centering
    \includegraphics[width=\linewidth]{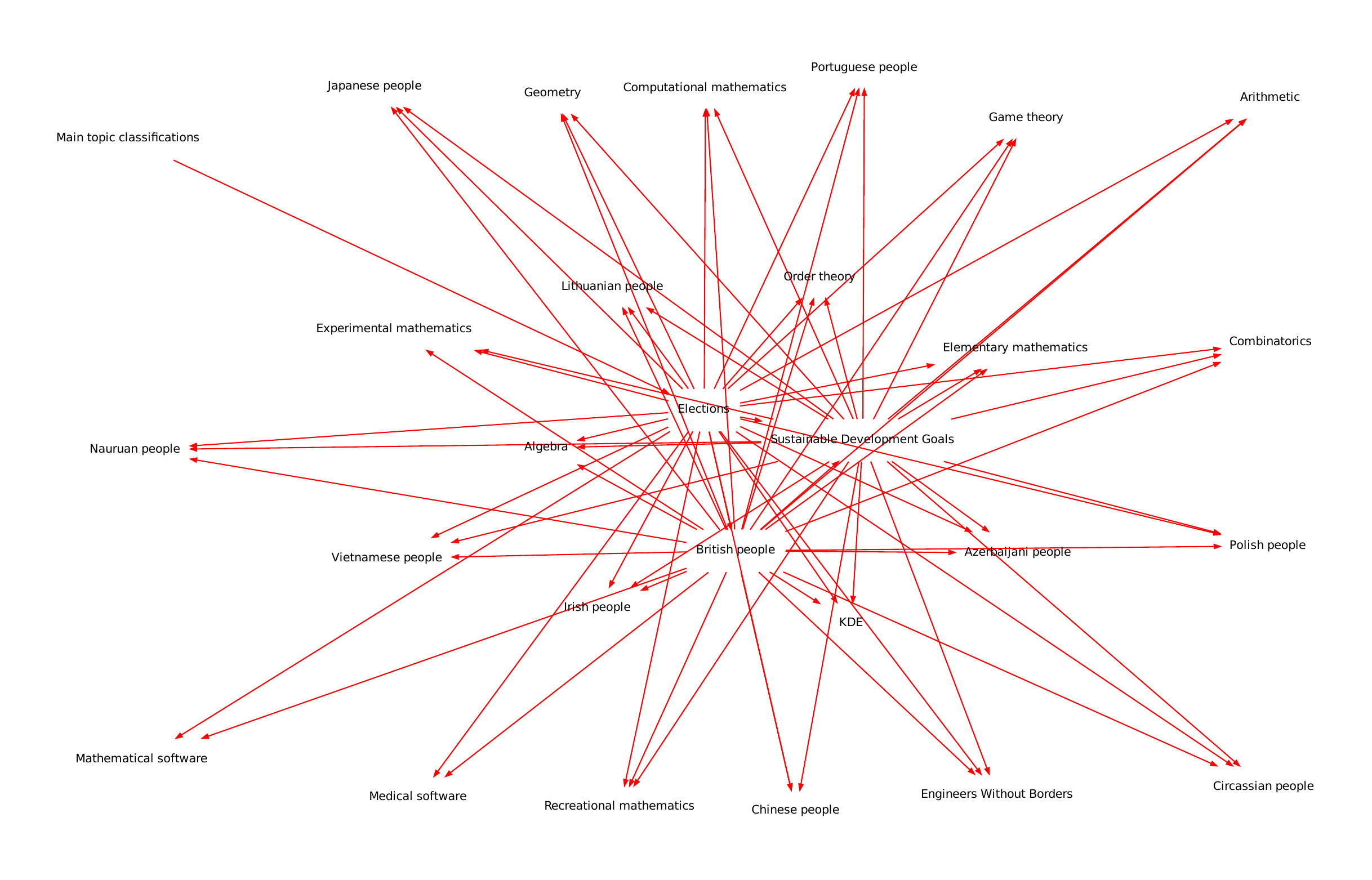}
    \caption{Elections}
    \end{subfigure}
    \begin{subfigure}{0.8\textwidth}
    \centering
    \includegraphics[width=\linewidth]{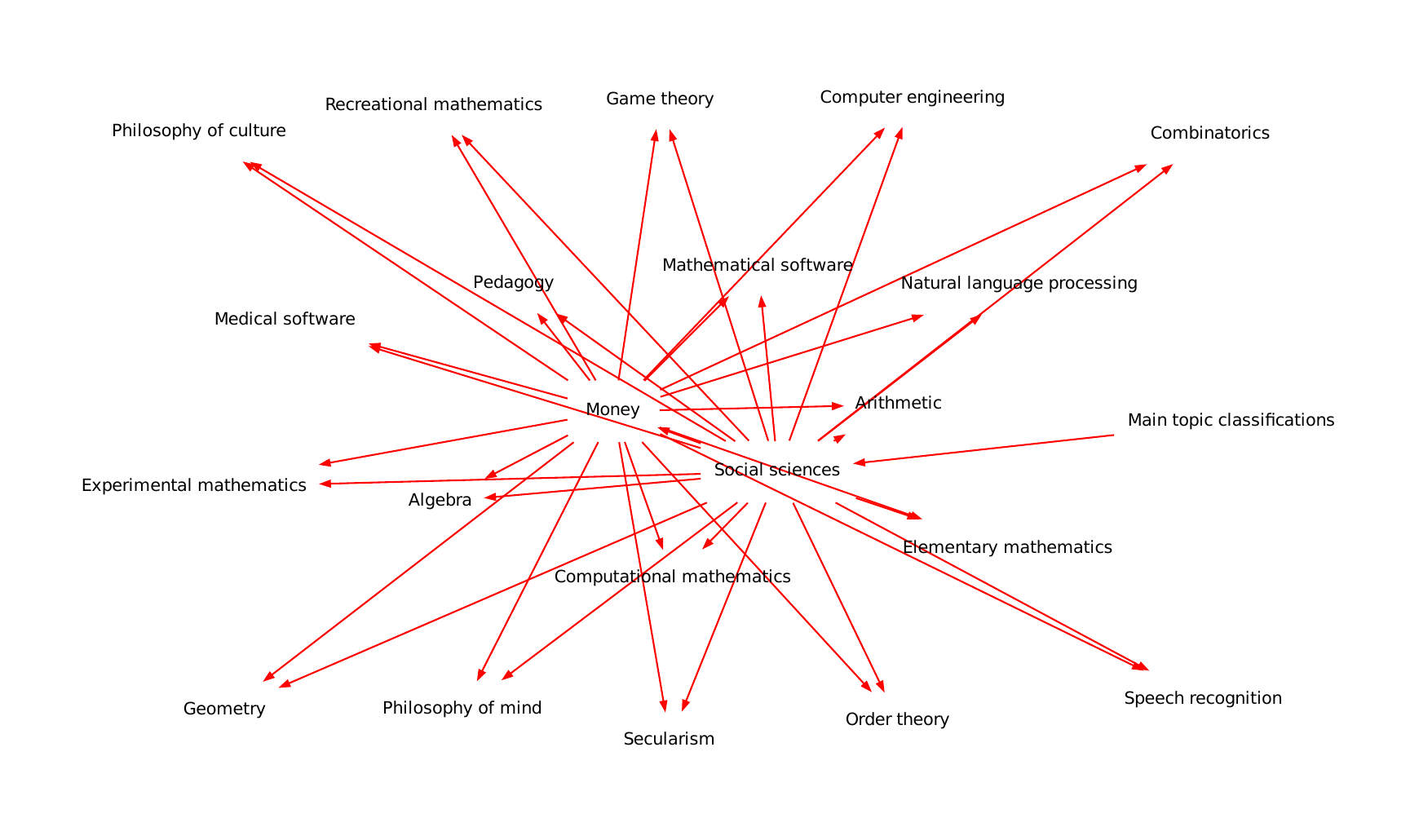}
    \caption{Money}
    \end{subfigure}
    \begin{subfigure}{0.8\textwidth}
    \centering
    \includegraphics[width=\linewidth]{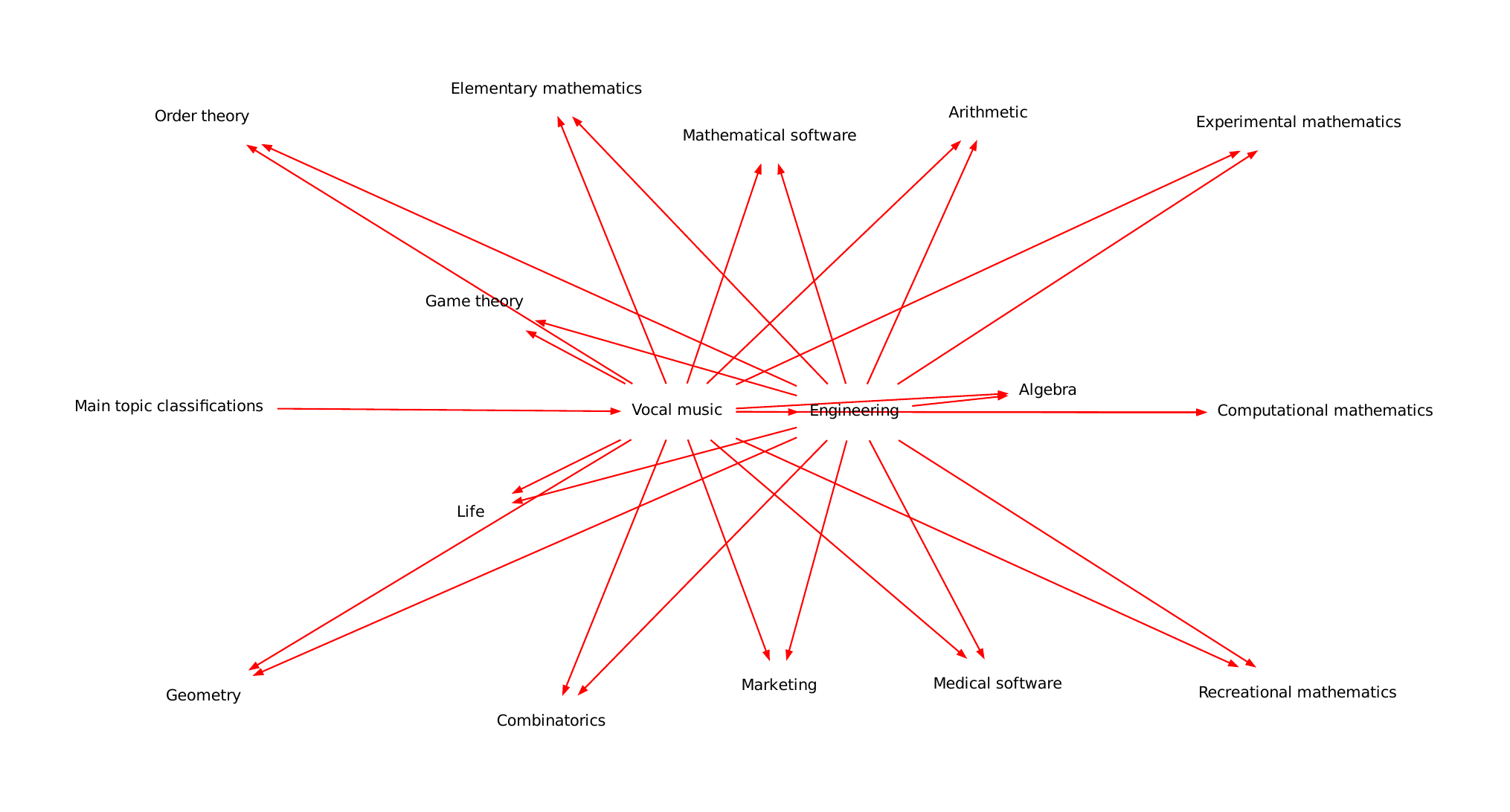}
    \caption{Vocal music}
    \end{subfigure}
    \caption{Sub-ontologies for Wikipedia generated by REBEL, centred on various topics.}
\end{figure}

\begin{figure}[H]
    \centering
    \begin{subfigure}{1.0\textwidth}
    \centering
    \includegraphics[width=\linewidth]{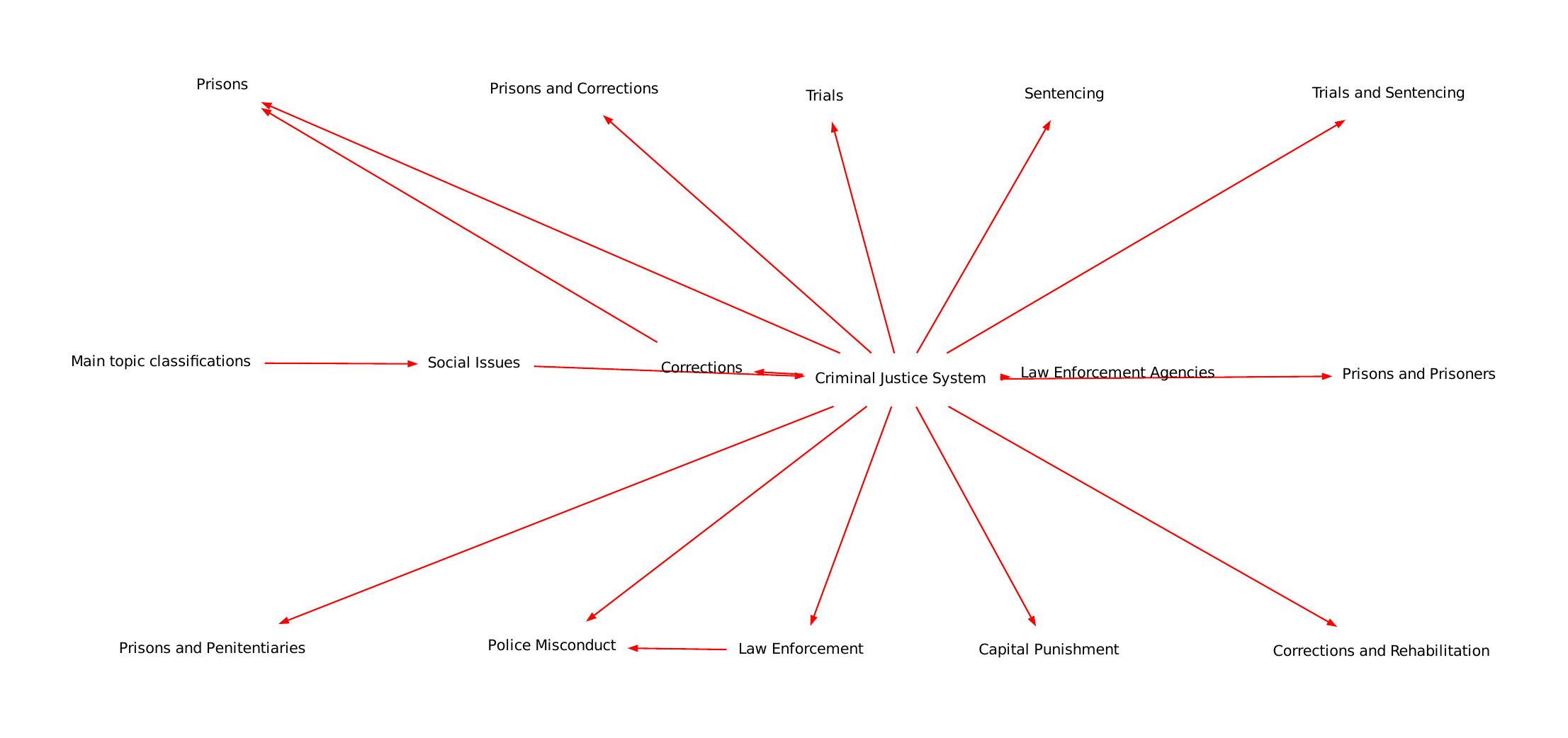}
    \caption{Criminal Justice System}
    \end{subfigure}
    \begin{subfigure}{1.0\textwidth}
    \centering
    \includegraphics[width=\linewidth]{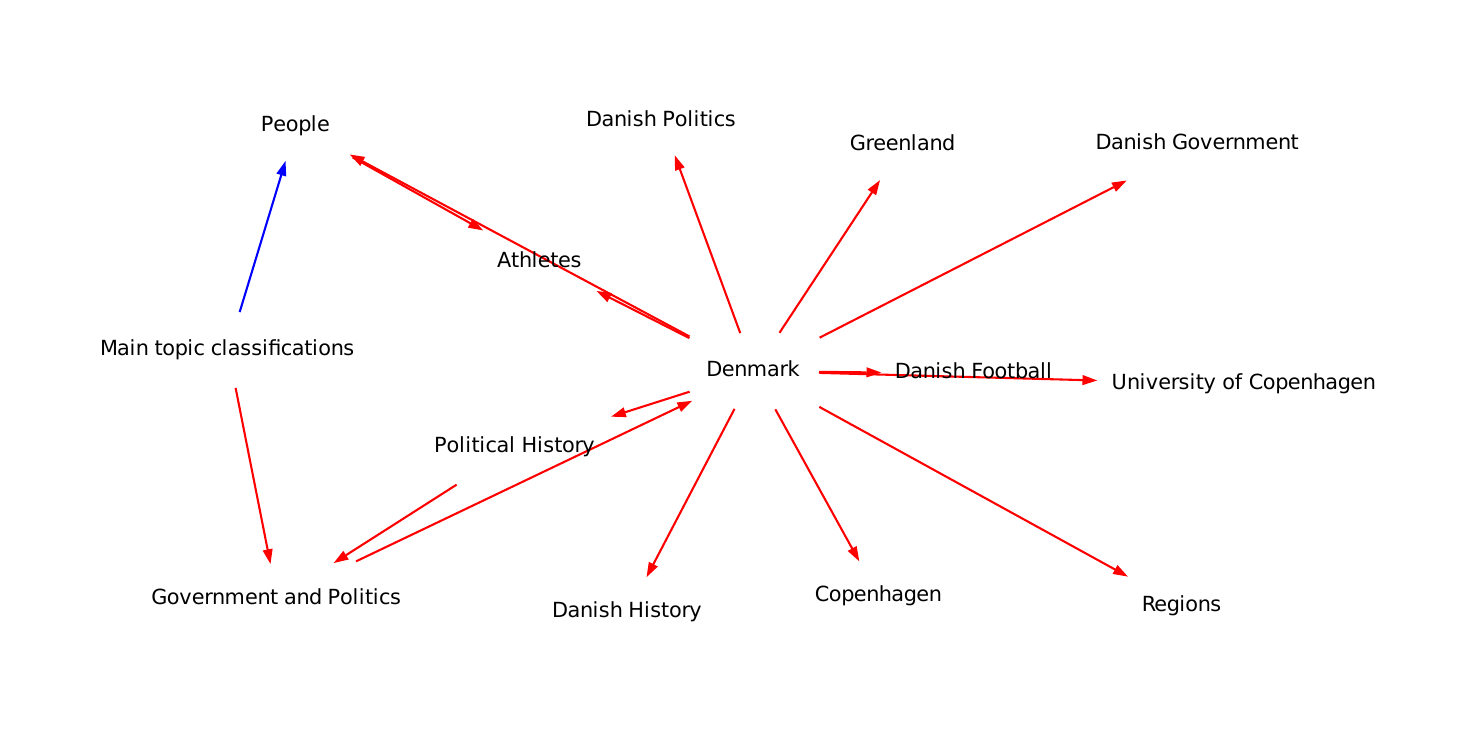}
    \caption{Denmark}
    \end{subfigure}
    \begin{subfigure}{1.0\textwidth}
    \centering
    \includegraphics[width=\linewidth]{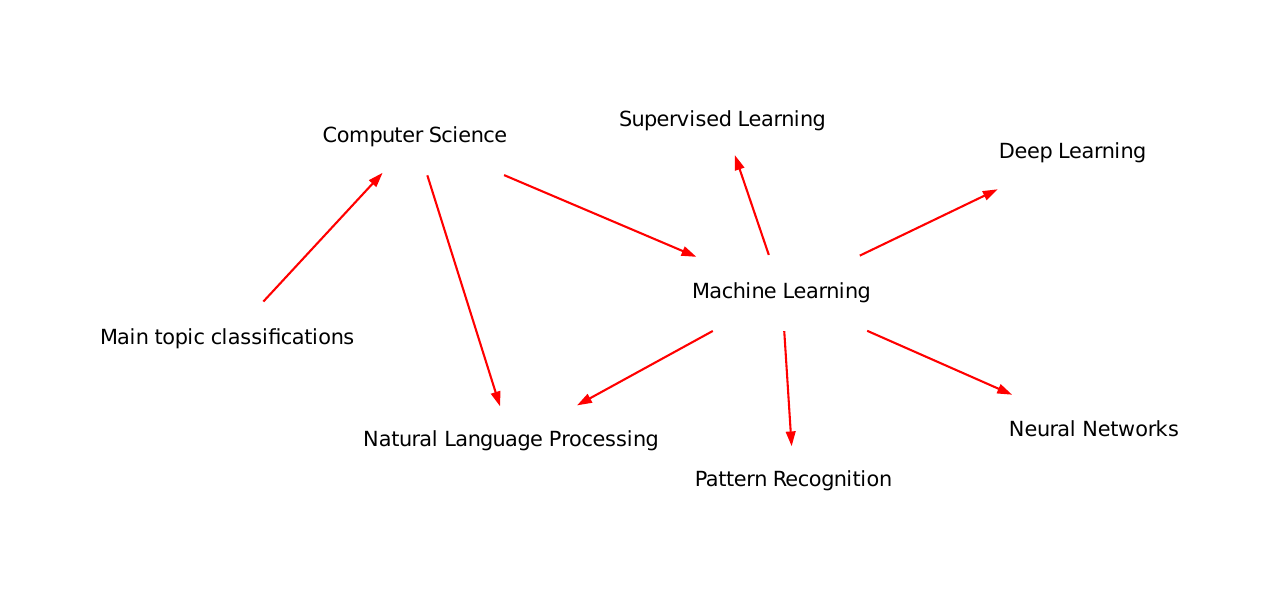}
    \caption{Machine Learning}
    \end{subfigure}
    \caption{Sub-ontologies for Wikipedia generated by Zero-shot, centred on various topics.}
\end{figure}

\begin{figure}[H]
    \centering
    \begin{subfigure}{1.0\textwidth}
    \centering
    \includegraphics[width=\linewidth]{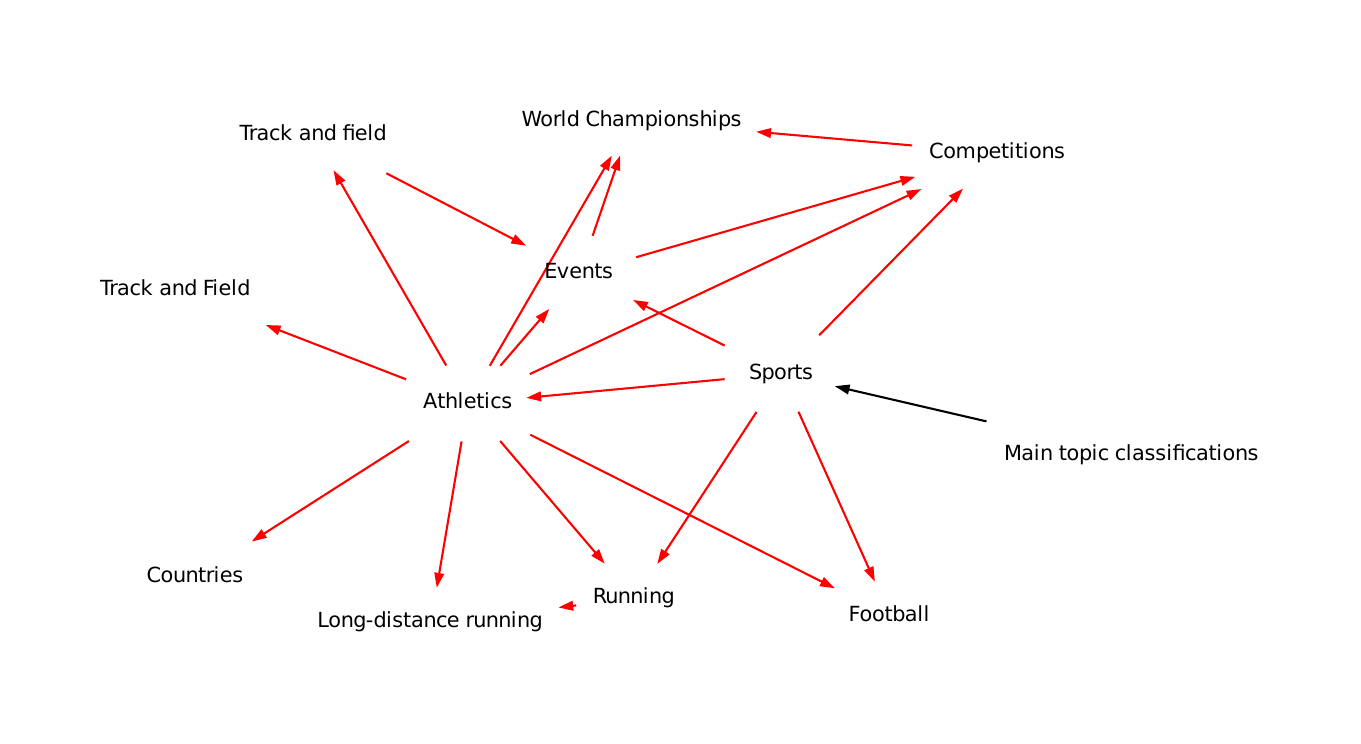}
    \caption{Athletics}
    \end{subfigure}
    \begin{subfigure}{1.0\textwidth}
    \centering
    \includegraphics[width=\linewidth]{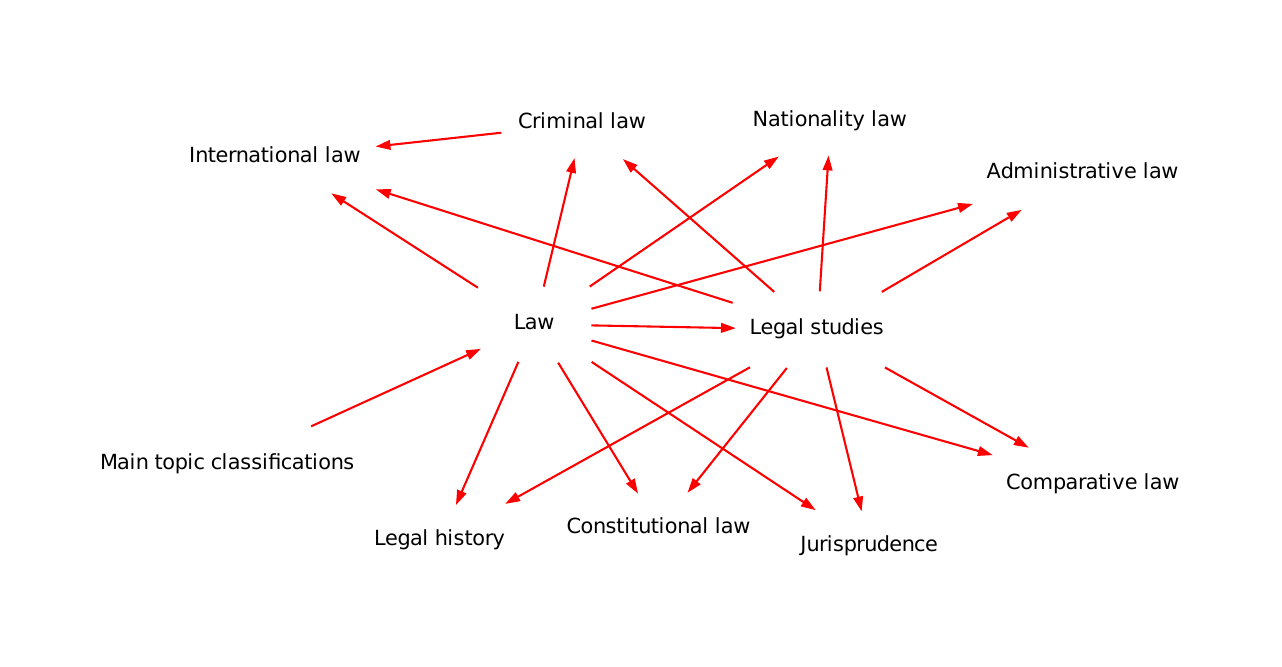}
    \caption{Legal studies}
    \end{subfigure}
    \begin{subfigure}{1.0\textwidth}
    \centering
    \includegraphics[width=\linewidth]{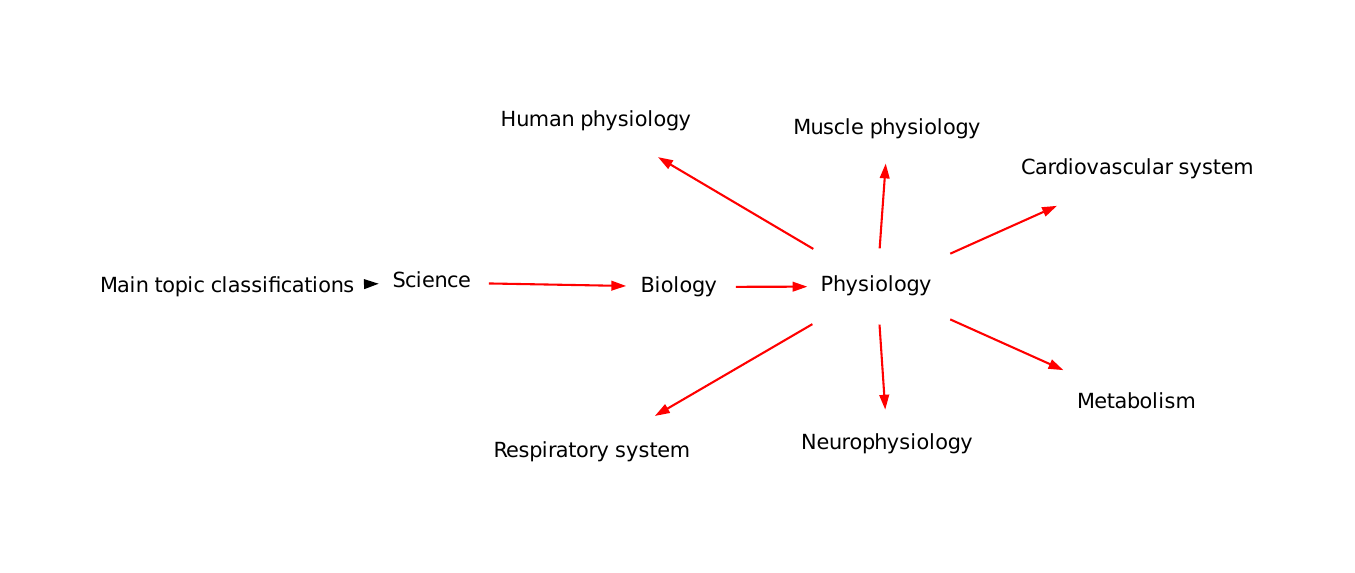}
    \caption{Physiology}
    \end{subfigure}
    \caption{Sub-ontologies for Wikipedia generated by One-shot, centred on various topics.}
\end{figure}

\begin{figure}[H]
    \centering
    \begin{subfigure}{0.9\textwidth}
    \centering
    \includegraphics[width=\linewidth]{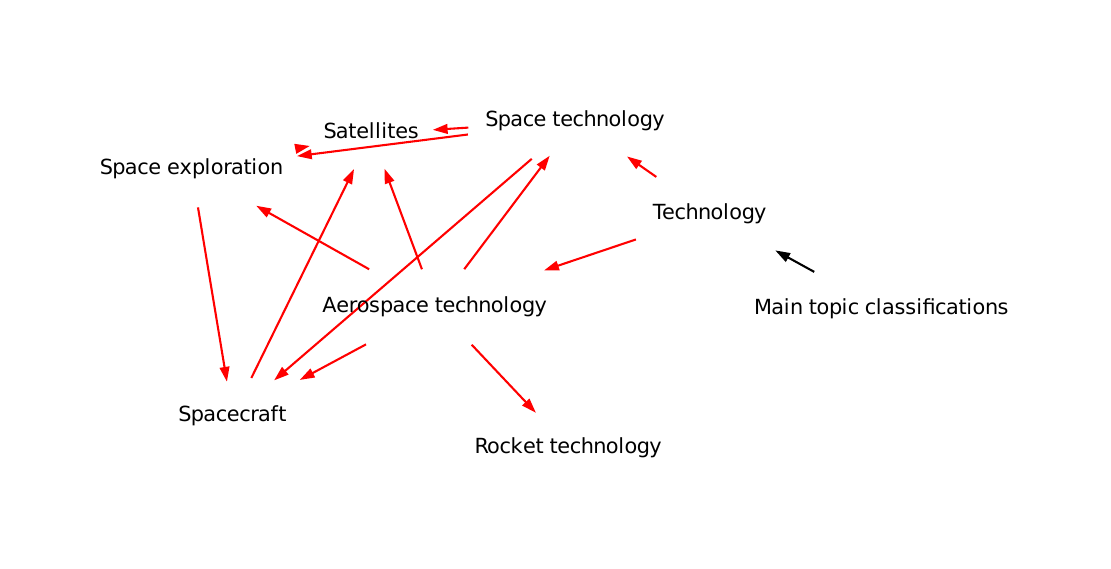}
    \caption{Aerospace technology}
    \end{subfigure}
    \begin{subfigure}{1.0\textwidth}
    \centering
    \includegraphics[width=\linewidth]{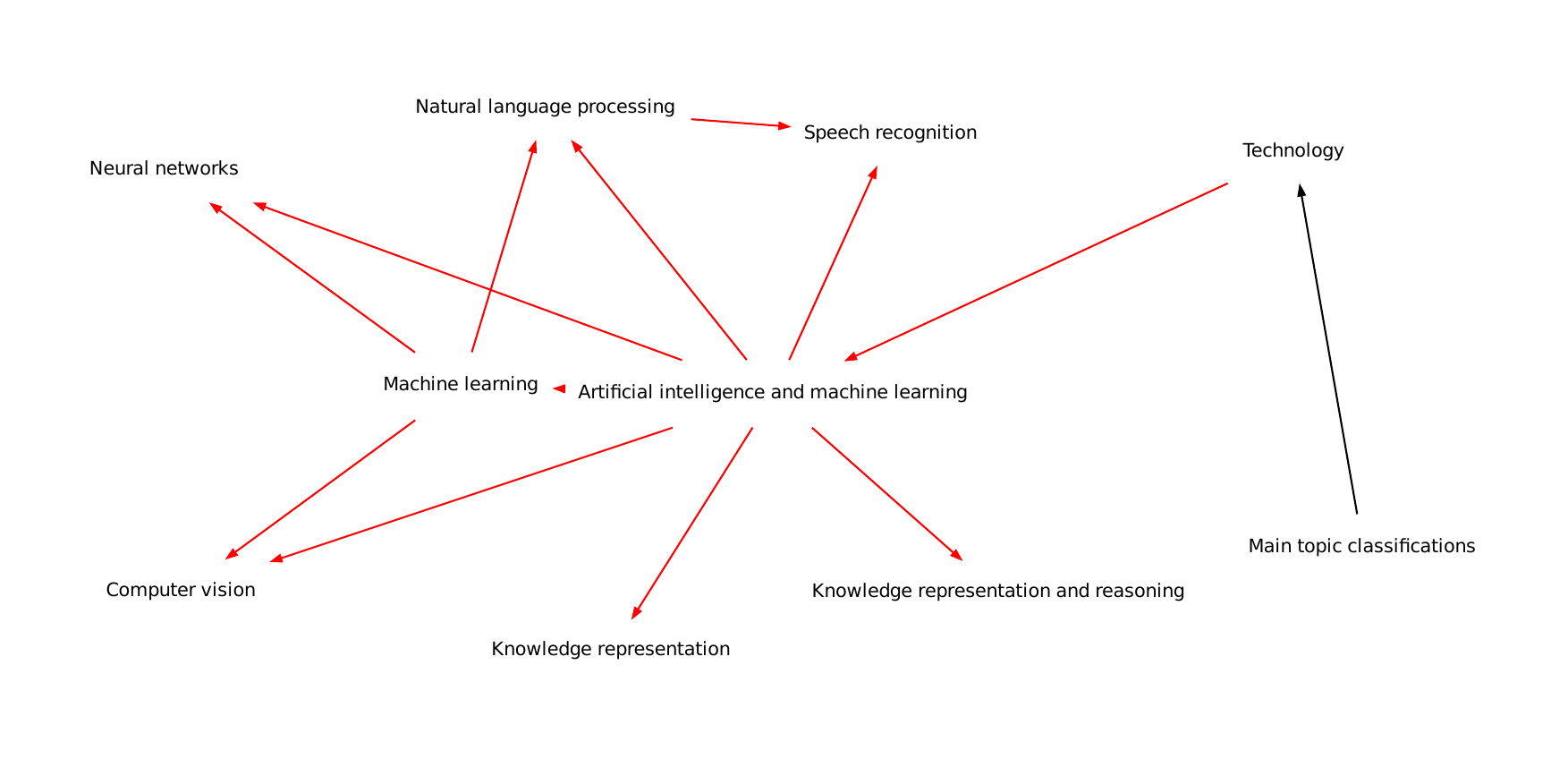}
    \caption{Artificial intelligence and machine learning}
    \end{subfigure}
    \begin{subfigure}{1.0\textwidth}
    \centering
    \includegraphics[width=\linewidth]{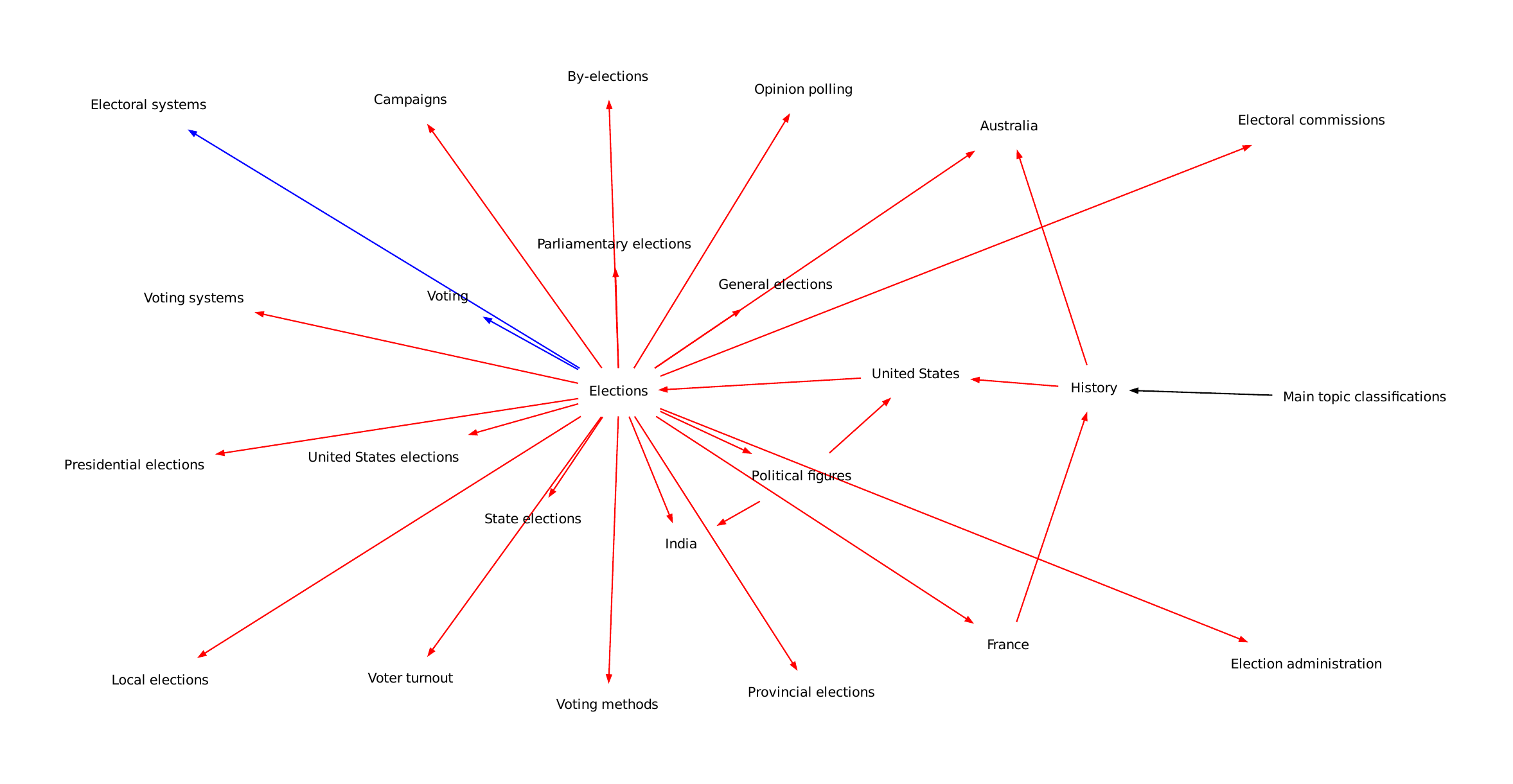}
    \caption{Elections}
    \label{fig:3shot-wiki-samples-election}
    \end{subfigure}
    \caption{Sub-ontologies for Wikipedia generated by Three-shot, centred on various topics.}
\end{figure}

\subsubsection{arXiv}  \label{appendix:viz-arxiv}

\begin{figure}[h]
    \centering
    \includegraphics[width=\linewidth]{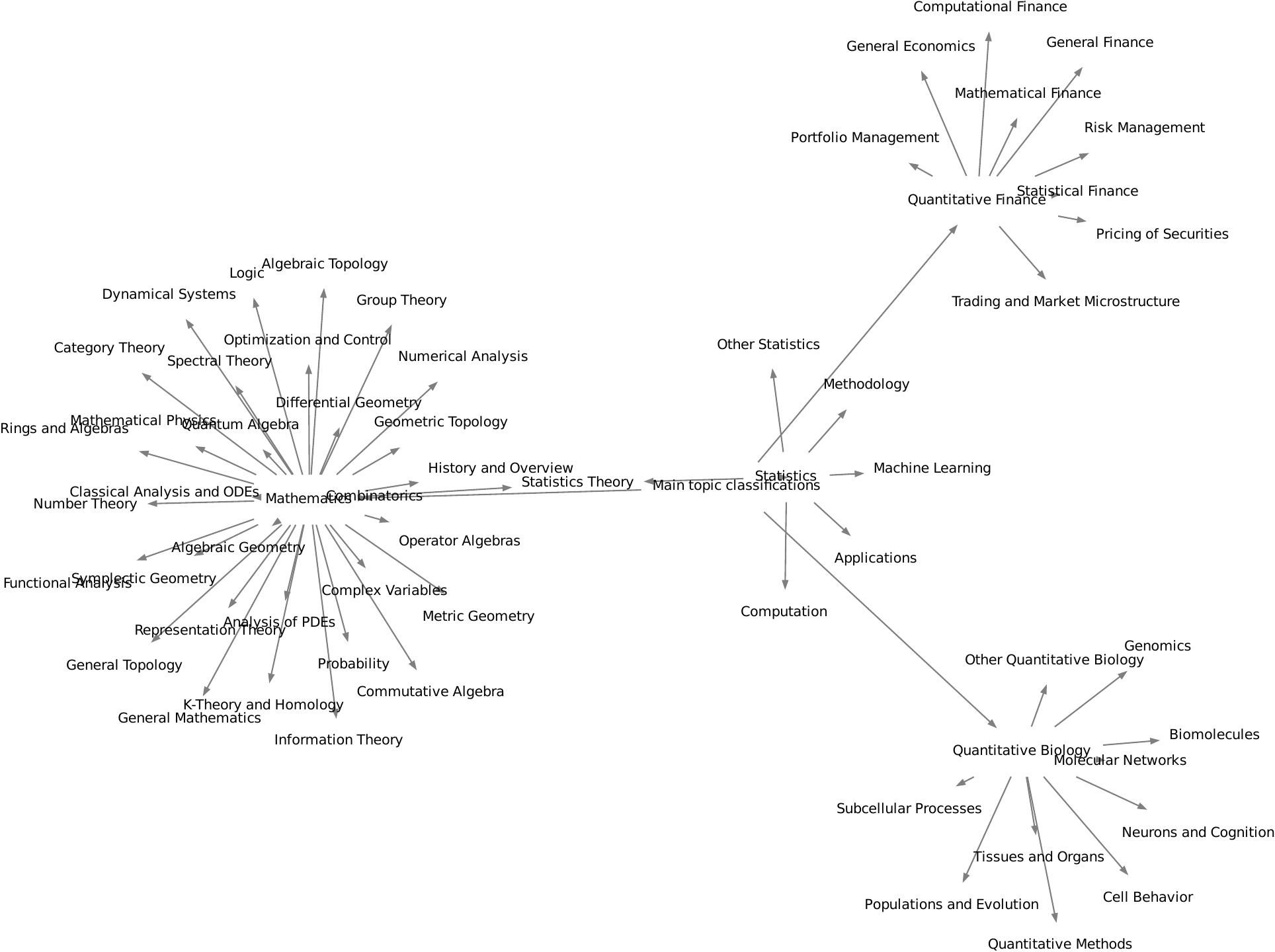}
    \caption{Ground truth test split ontology for arXiv}
\end{figure}


\begin{figure}[h]
    \centering
    \includegraphics[width=\linewidth]{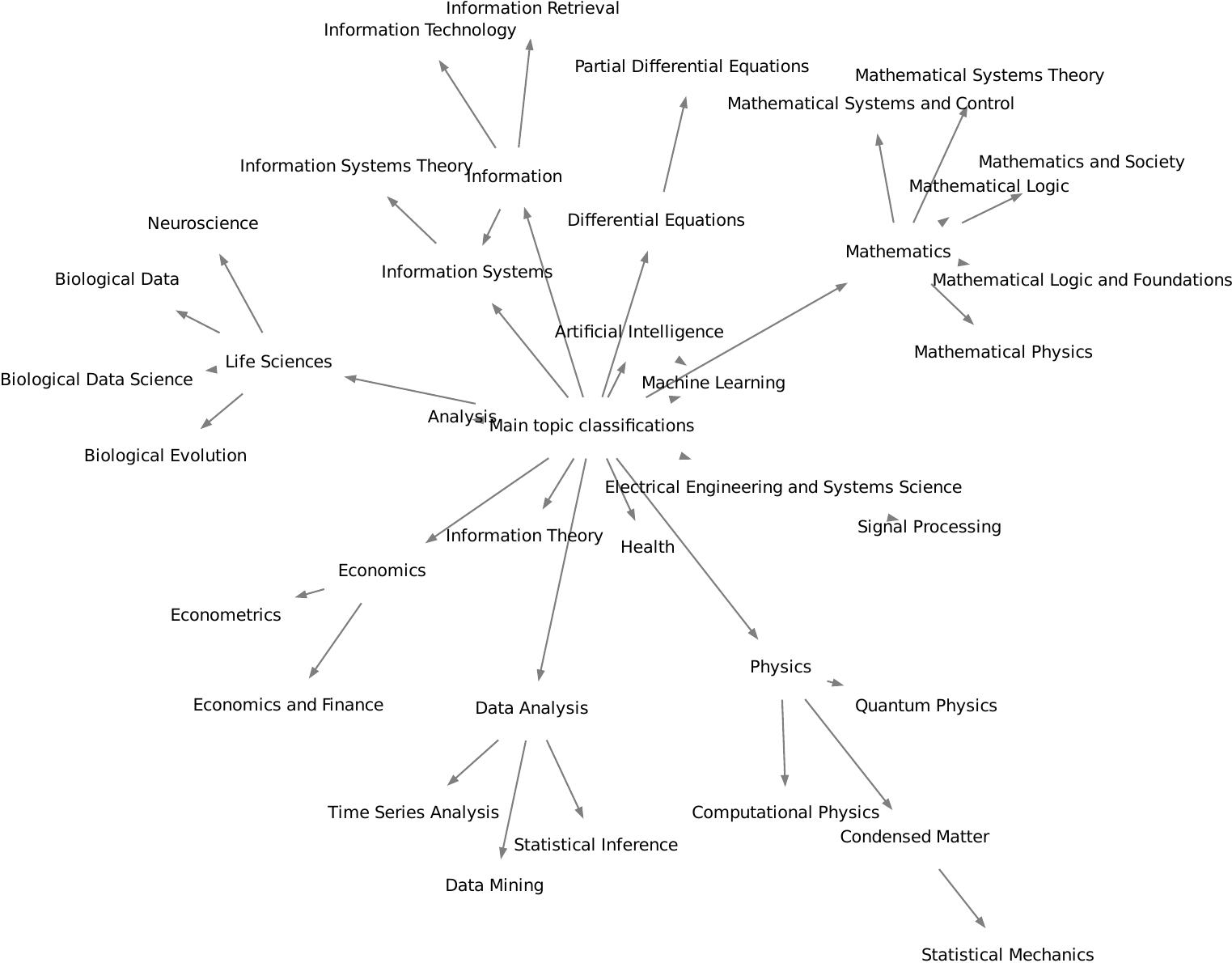}
    \caption{Ontology for arXiv generated by \name}
    \label{fig:ollm-arxiv}
\end{figure}


\begin{figure}[h]
    \centering
    \includegraphics[width=\linewidth]{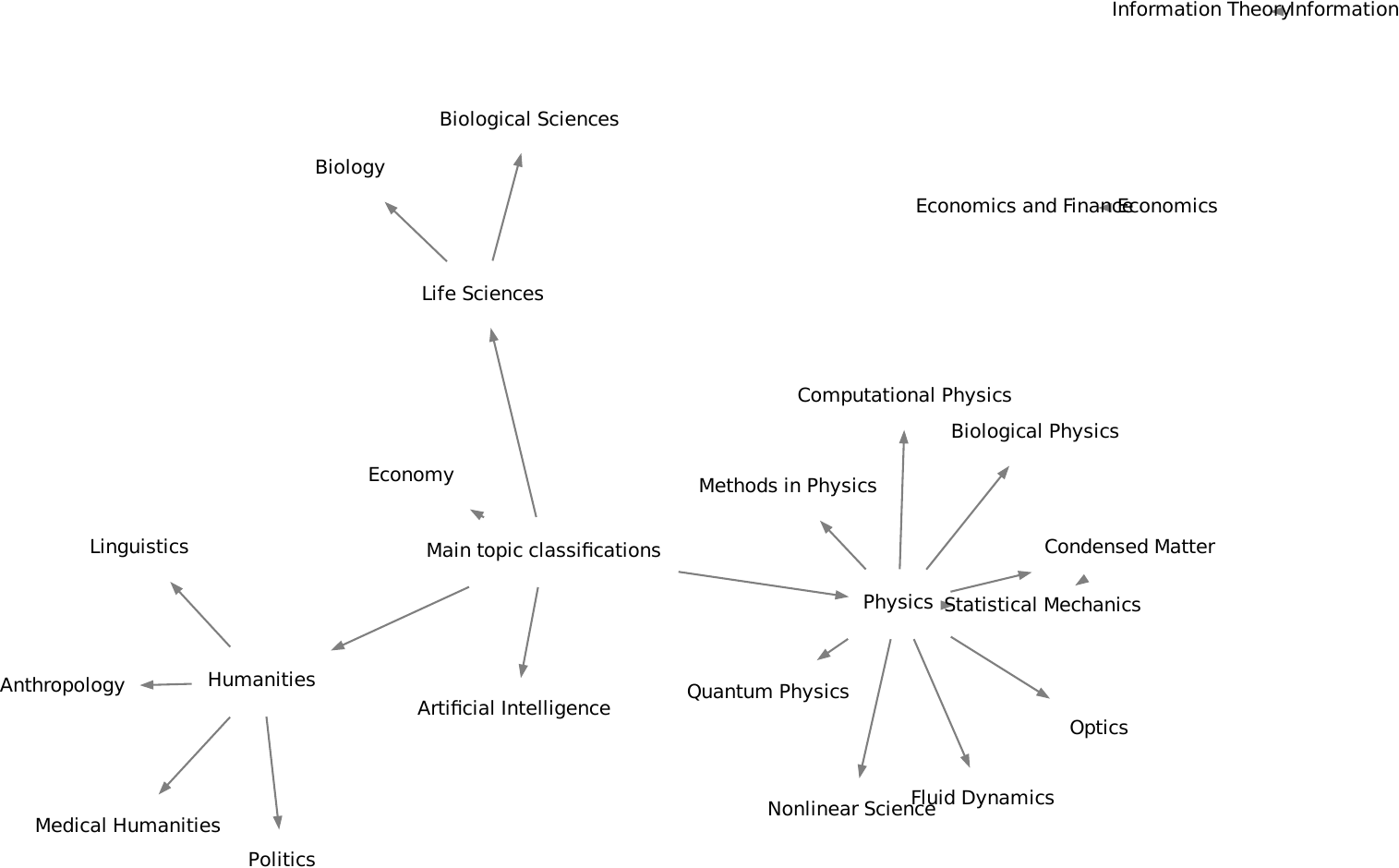}
    \caption{Ontology for arXiv generated by Finetune}
\end{figure}


\begin{figure}[h]
    \centering
    \includegraphics[width=\linewidth]{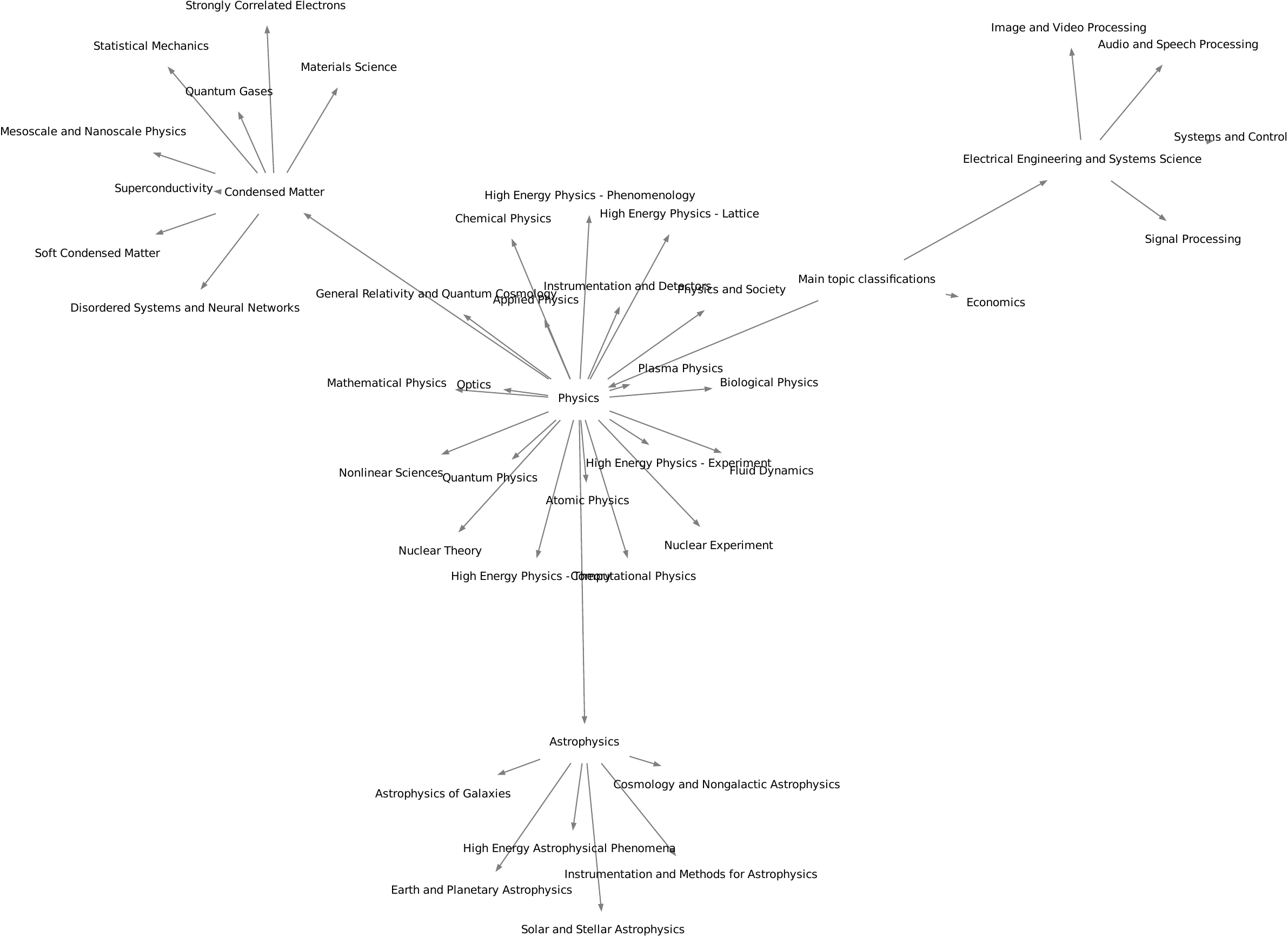}
    \caption{Ontology for arXiv generated by Memorisation}
\end{figure}


\begin{figure}[h]
    \centering
    \includegraphics[width=\linewidth]{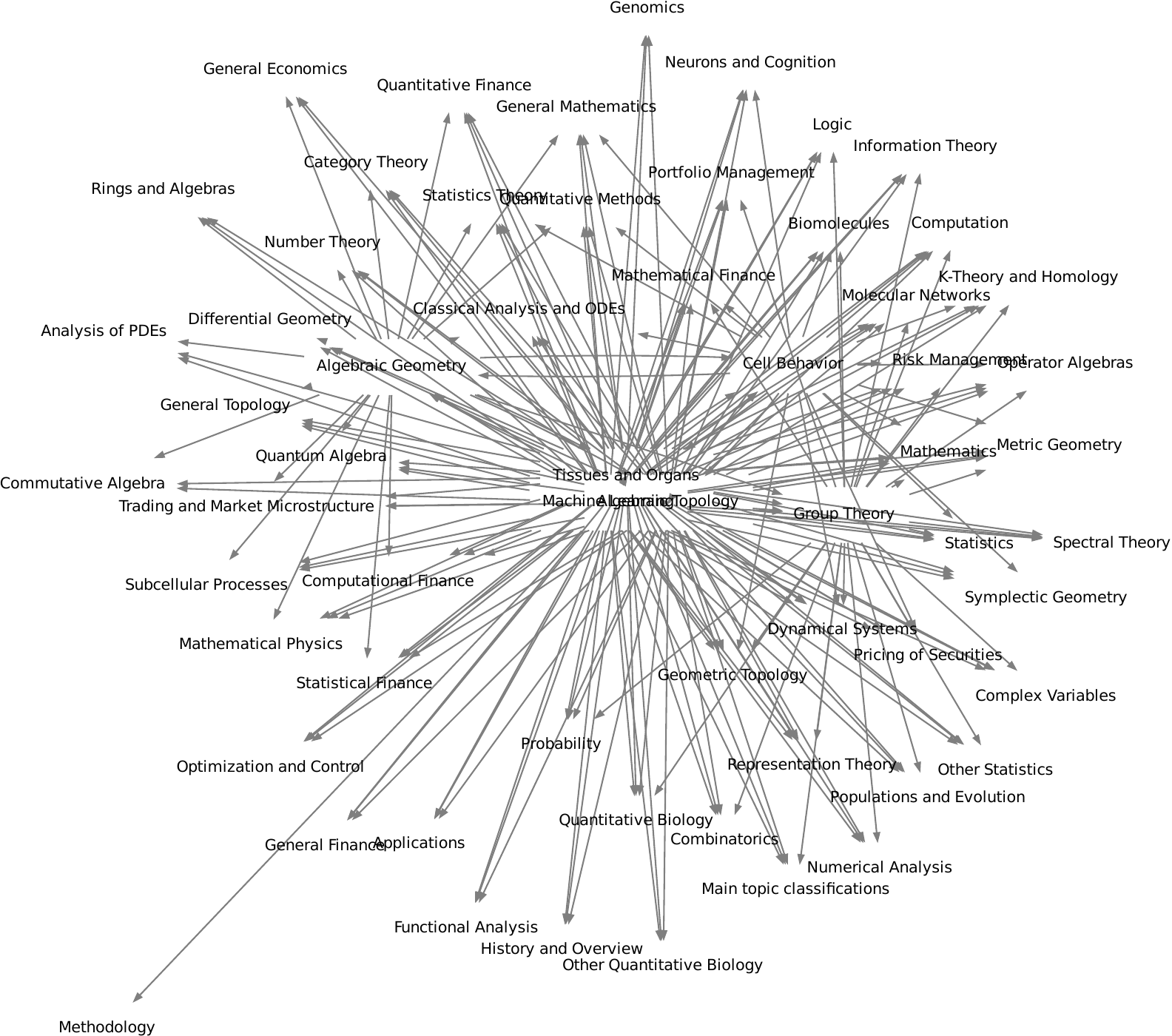}
    \caption{Ontology for arXiv generated by Hearst}
    \label{fig:hearst-arxiv}
\end{figure}


\begin{figure}[h]
    \centering
    \includegraphics[width=\linewidth]{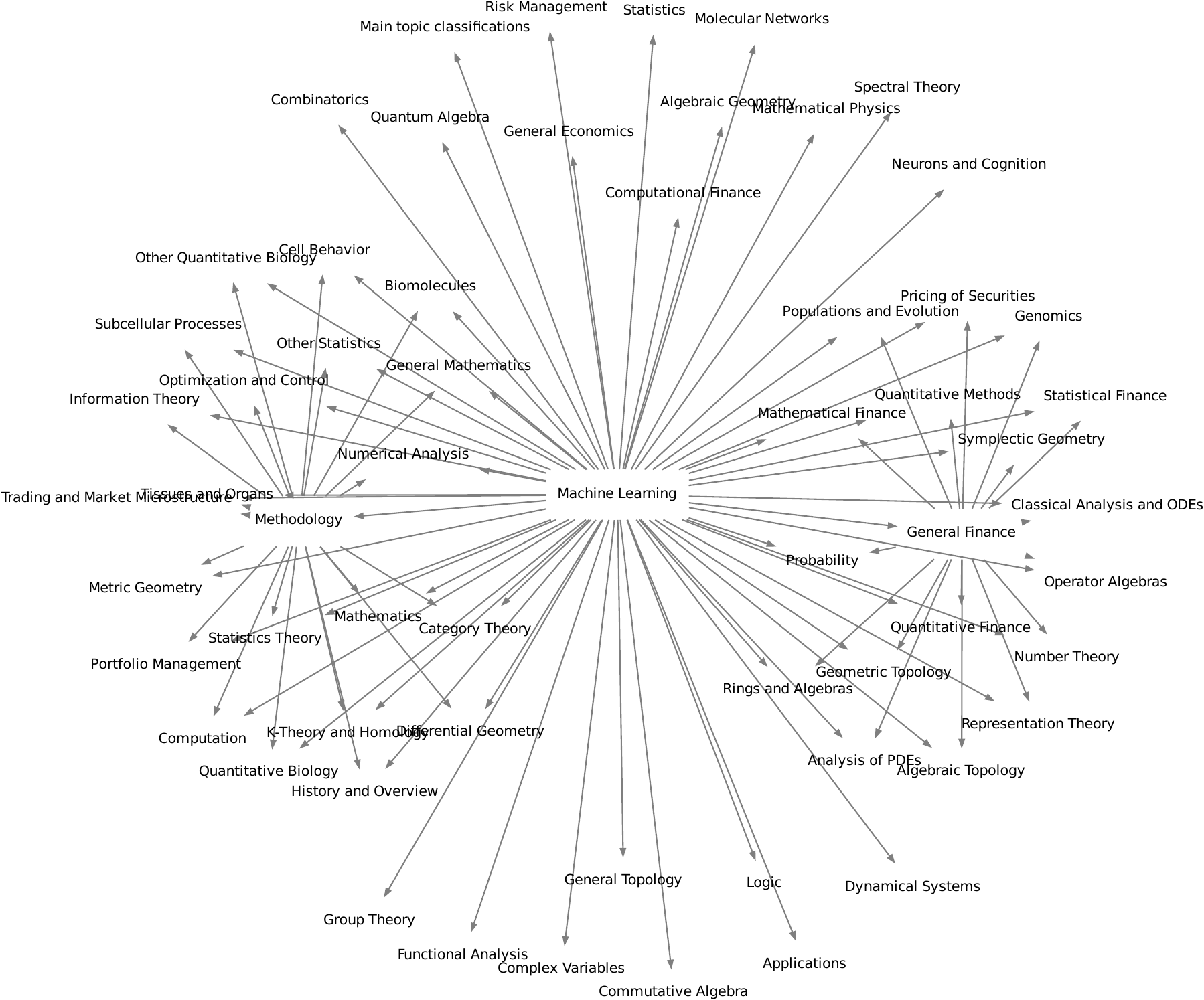}
    \caption{Ontology for arXiv generated by REBEL}
    \label{fig:rebel-arxiv}
\end{figure}


\begin{figure}[h]
    \centering
    \includegraphics[width=\linewidth]{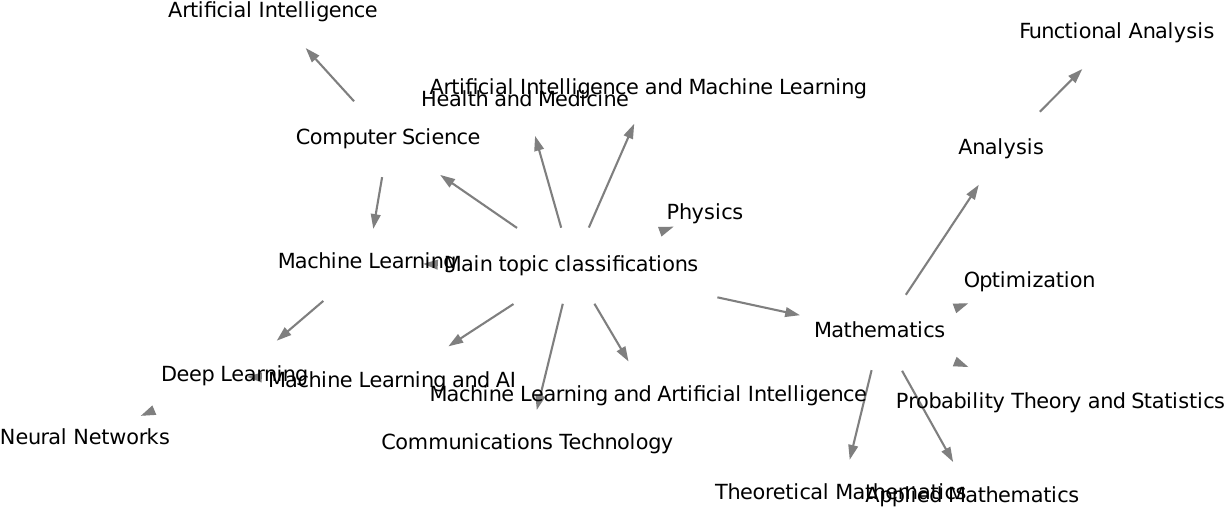}
    \caption{Ontology for arXiv generated by Zero-shot}
\end{figure}

\begin{figure}[h]
    \centering
    \includegraphics[width=\linewidth]{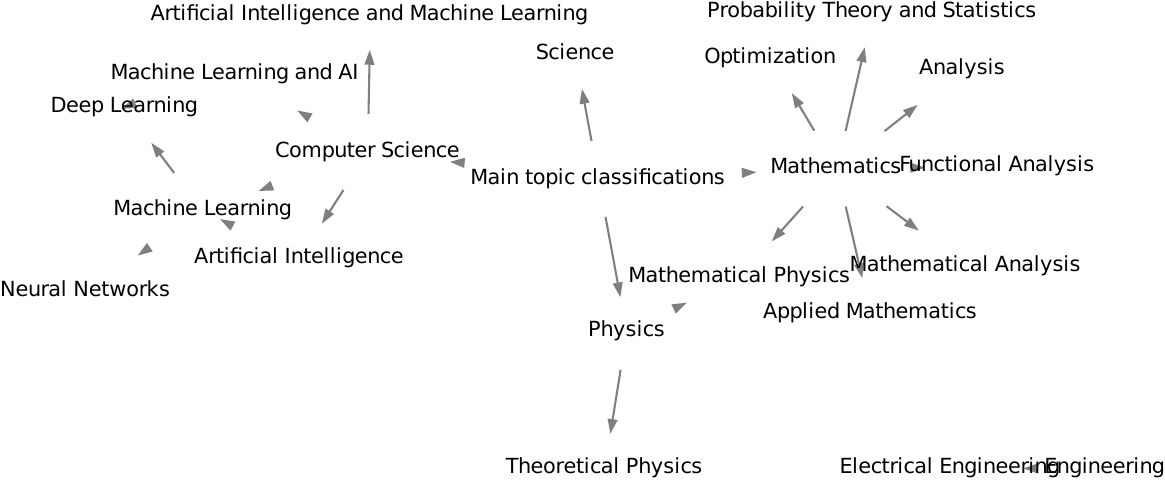}
    \caption{Ontology for arXiv generated by One-shot}
\end{figure}

\begin{figure}[h]
    \centering
    \includegraphics[width=\linewidth]{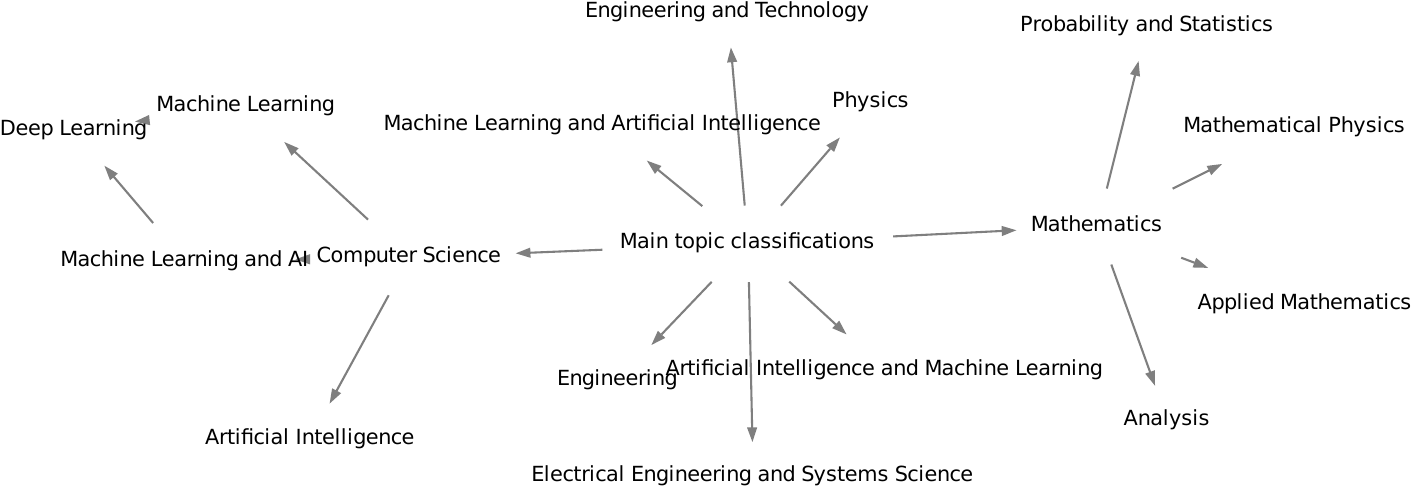}
    \caption{Ontology for arXiv generated by Three-shot}
    \label{fig:3shot-arxiv}
\end{figure}


\newpage
\section*{NeurIPS Paper Checklist}

\begin{enumerate}

\item {\bf Claims}
    \item[] Question: Do the main claims made in the abstract and introduction accurately reflect the paper's contributions and scope?
    \item[] Answer: \answerYes{}
    \item[] Justification: The main claims of this paper are: 1. \name is an effective method for building ontologies from scratch, and 2. the evaluation metrics we introduce are robust and useful for gold standard evaluation of ontologies. We justify the first claim by demonstrating that \name outperforms our baseline methods on Wikipedia and arXiv according to our metrics. We justify the second claim by showing that an existing metric (Literal F1) can score the sub-optimal Memorisation solution highly, while our metrics are not subject to this issue. Our new metrics also suggest results that align with our qualitative analysis. 

\item {\bf Limitations}
    \item[] Question: Does the paper discuss the limitations of the work performed by the authors?
    \item[] Answer: \answerYes{}
    \item[] Justification: We discuss the limitations and the possible resolutions in \cref{sec:disccusion}. This includes the scope of the work, where we only study ontologies with concepts and taxonomic relations, the inability to guarantee taxonomic relation transitivity, and the inability to control for data leakage from the pretraining stage of the LLM base model.

\item {\bf Theory Assumptions and Proofs}
    \item[] Question: For each theoretical result, does the paper provide the full set of assumptions and a complete (and correct) proof?
    \item[] Answer: \answerNA{}
    \item[] Justification: This paper does not include theoretical results.

    \item {\bf Experimental Result Reproducibility}
    \item[] Question: Does the paper fully disclose all the information needed to reproduce the main experimental results of the paper to the extent that it affects the main claims and/or conclusions of the paper (regardless of whether the code and data are provided or not)?
    \item[] Answer: \answerYes{}
    \item[] Justification: We describe the data collection procedure in \cref{sec:dataset} and the full experiment details in \cref{appendix:exp-details}. We also include the code and dataset in the supplementary material. 

\item {\bf Open access to data and code}
    \item[] Question: Does the paper provide open access to the data and code, with sufficient instructions to faithfully reproduce the main experimental results, as described in supplemental material?
    \item[] Answer: \answerYes{}
    \item[] Justification: The code and data used for this project are provided in the supplementary material. The code includes a README which details the steps for reproducing our results. 

\item {\bf Experimental Setting/Details}
    \item[] Question: Does the paper specify all the training and test details (e.g., data splits, hyperparameters, how they were chosen, type of optimizer, etc.) necessary to understand the results?
    \item[] Answer: \answerYes{}
    \item[] Justification: We give all the experimental details in \cref{appendix:exp-details}. 

\item {\bf Experiment Statistical Significance}
    \item[] Question: Does the paper report error bars suitably and correctly defined or other appropriate information about the statistical significance of the experiments?
    \item[] Answer: \answerNo{}
    \item[] Justification: We did not perform repeated experiments due to compute constraints, therefore there are no error bars. 

\item {\bf Experiments Compute Resources}
    \item[] Question: For each experiment, does the paper provide sufficient information on the computer resources (type of compute workers, memory, time of execution) needed to reproduce the experiments?
    \item[] Answer: \answerYes{}
    \item[] Justification: We describe the compute requirements for each experiment in \cref{appendix:exp-details}. 
    
\item {\bf Code Of Ethics}
    \item[] Question: Does the research conducted in the paper conform, in every respect, with the NeurIPS Code of Ethics \url{https://neurips.cc/public/EthicsGuidelines}?
    \item[] Answer: \answerYes{}
    \item[] Justification: This paper does not involve human subjects and does not use data that is not already in the public domain. We clearly describe our data collection procedure in \cref{sec:dataset}. Our method is highly specialised in building ontologies and solving related tasks, thus having minimal societal impact or any potentially harmful consequences. 

\item {\bf Broader Impacts}
    \item[] Question: Does the paper discuss both potential positive societal impacts and negative societal impacts of the work performed?
    \item[] Answer: \answerNA{}
    \item[] Justification: Our method is highly specialised in building ontologies and solving related tasks only. We do not expect our work to have a wider impact than improving the quality of existing or new ontologies.
    
\item {\bf Safeguards}
    \item[] Question: Does the paper describe safeguards that have been put in place for responsible release of data or models that have a high risk for misuse (e.g., pretrained language models, image generators, or scraped datasets)?
    \item[] Answer: \answerNA{}
    \item[] Justification: This paper poses no such risks. The trained model is specific to building ontologies only.

\item {\bf Licenses for existing assets}
    \item[] Question: Are the creators or original owners of assets (e.g., code, data, models), used in the paper, properly credited and are the license and terms of use explicitly mentioned and properly respected?
    \item[] Answer: \answerYes{}
    \item[] Justification: The two datasets collected in this project use data from Wikipedia and arXiv, which are in the public domain under the CC BY-SA 4.0 and CC0 1.0 Deed licenses respectively. The REBEL-large model is available under CC BY-NC-SA 4.0 license.

\item {\bf New Assets}
    \item[] Question: Are new assets introduced in the paper well documented and is the documentation provided alongside the assets?
    \item[] Answer: \answerYes{}
    \item[] Justification: We release the dataset and code for this paper. The data collection procedure is clearly described in \cref{sec:dataset}. The training procedure is clearly described in \cref{appendix:exp-details}. 

\item {\bf Crowdsourcing and Research with Human Subjects}
    \item[] Question: For crowdsourcing experiments and research with human subjects, does the paper include the full text of instructions given to participants and screenshots, if applicable, as well as details about compensation (if any)? 
    \item[] Answer: \answerNA{}
    \item[] Justification: This paper does not involve crowdsourcing nor research with human subjects.

\item {\bf Institutional Review Board (IRB) Approvals or Equivalent for Research with Human Subjects}
    \item[] Question: Does the paper describe potential risks incurred by study participants, whether such risks were disclosed to the subjects, and whether Institutional Review Board (IRB) approvals (or an equivalent approval/review based on the requirements of your country or institution) were obtained?
    \item[] Answer: \answerNA{}
    \item[] Justification: This paper does not involve crowdsourcing nor research with human subjects.

\end{enumerate}

\end{document}